\pdfoutput=1

\PassOptionsToPackage{table,xcdraw}{xcolor}

\documentclass[11pt]{article}

\usepackage[final]{acl}

\usepackage{times}
\usepackage{latexsym}

\usepackage[T1]{fontenc}

\usepackage[utf8]{inputenc}

\usepackage{microtype}

\usepackage{inconsolata}

\usepackage{booktabs}
\usepackage{multirow}
\usepackage{multicol}
\usepackage{makecell}
\usepackage{enumitem}
\usepackage{balance}
\usepackage{graphicx}
\usepackage{subfigure}
\usepackage{amsmath}
\usepackage{amsthm}
\usepackage{amssymb}
\usepackage{amsfonts}
\usepackage{hyperref}

\usepackage{booktabs}
\usepackage{soul}
\usepackage{todonotes}
\usepackage{xspace}

\newcommand{\eval}{\textsc{Dict}\xspace}
\newcommand{\model}{\textsc{Traver}\xspace}

\title{Training Turn-by-Turn Verifiers for Dialogue Tutoring Agents: \\ The Curious Case of LLMs as Your Coding Tutors}

\author{%
    Jian Wang$^{1,2}$\thanks{\ \ This work was conducted while Jian Wang was a visiting PhD student at the University of Michigan.} \quad  Yinpei Dai$^{2}$  \quad  Yichi Zhang$^2$  \\
    {\bf Ziqiao Ma$^2$  \qquad  Wenjie Li$^1$  \qquad  Joyce Chai$^2$}  \\
    $^1$The Hong Kong Polytechnic University \quad
    $^2$University of Michigan  \\
    \texttt{jian-dylan.wang@connect.polyu.hk  ~~ cswjli@comp.polyu.edu.hk} \\
    \texttt{\{daiyp,zhangyic,marstin,chaijy\}@umich.edu}
}

\begin{document}
\maketitle

\begin{abstract}

Intelligent tutoring agents powered by large language models (LLMs) have been increasingly explored to deliver personalized knowledge in areas such as language learning and science education. However, their capabilities in guiding users to solve complex real-world tasks remain underexplored. To address this limitation, in this work, we focus on \textit{coding tutoring}, a challenging problem that requires tutors to proactively guide students towards completing predefined coding tasks. We propose a novel agent workflow, \textbf{Tra}ce-and-\textbf{Ver}ify (\textbf{\model}), which combines knowledge tracing to estimate a student's knowledge state and turn-by-turn verification to ensure effective guidance toward task completion. We introduce \textbf{\eval}, an automatic evaluation protocol that assesses tutor agents using controlled student simulation and code generation tests. Extensive experiments reveal the challenges of coding tutoring and demonstrate that \textsc{Traver} achieves a significantly higher success rate. Although we use code tutoring as an example in this paper, our approach can be extended beyond coding, providing valuable insights into advancing tutoring agents for human task learning.\footnote{Code and data are available at \url{https://github.com/iwangjian/Coding-Tutor}.}

\end{abstract}
\section{Introduction}

Tutoring has long been recognized as one of the most effective methods for enhancing human learning outcomes and addressing educational disparities~\citep{hill2005effects}. 
By providing personalized guidance to students, intelligent tutoring systems (ITS) have proven to be nearly as effective as human tutors in fostering deep understanding and skill acquisition, with research showing comparable learning gains~\citep{vanlehn2011relative,rus2013recent}.
More recently, the advancement of large language models (LLMs) has offered unprecedented opportunities to replicate these benefits in tutoring agents~\citep{dan2023educhat,jin2024teach,chen2024empowering}, unlocking the enormous potential to solve knowledge-intensive tasks such as answering complex questions or clarifying concepts.

\begin{figure}[t!]
\centering
\includegraphics[width=1.0\linewidth]{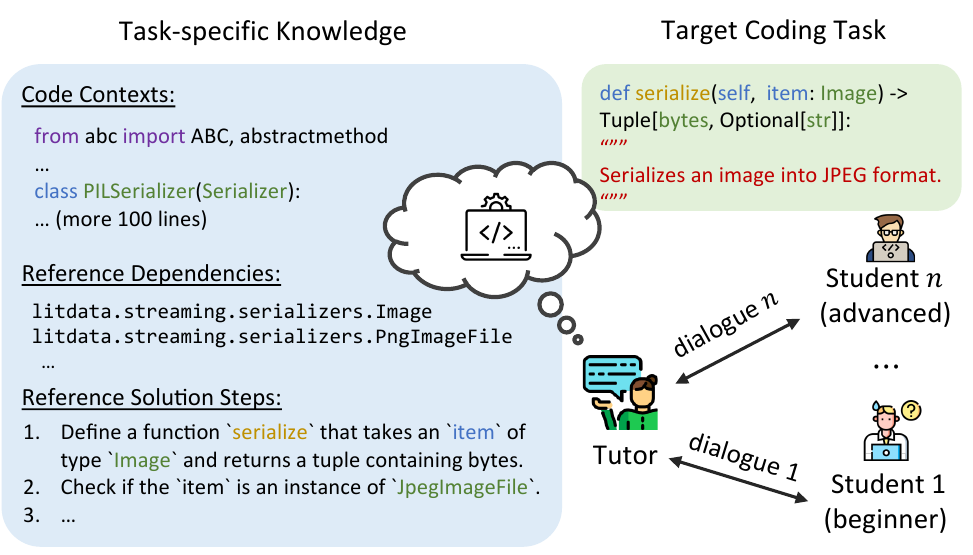}
\caption{An illustration of coding tutoring, where a tutor aims to proactively guide students toward completing a target coding task while adapting to students' varying levels of background knowledge. \vspace{-5pt}}
\label{fig:example}
\end{figure}

\begin{figure}[t!]
\centering
\includegraphics[width=1.0\linewidth]{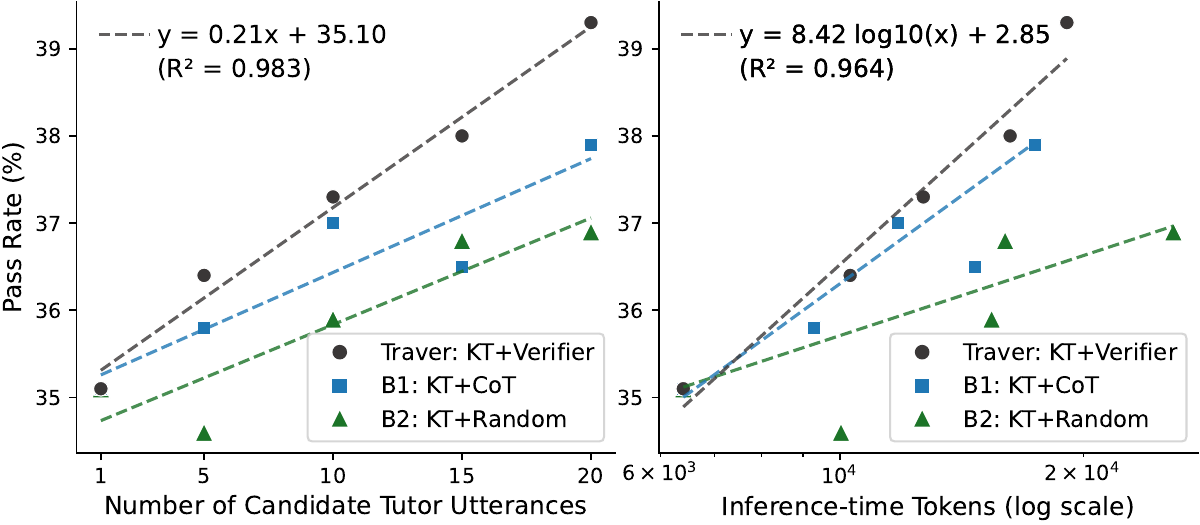}
\caption{\textsc{Traver} with the trained verifier shows inference-time scaling for coding tutoring (detailed in \S\ref{sec:scaling_analysis}). \textit{Left}: Performance vs. sampled candidate utterances per turn. \textit{Right}: Performance vs. total tokens consumed per tutoring session. \vspace{-15pt}}
\label{fig:scale}
\end{figure}

Previous research has extensively explored tutoring in \textit{knowledge delivery}, including language learning~\cite{swartz2012intelligent,stasaski-etal-2020-cima}, mathematical reasoning~\cite{demszky-hill-2023-ncte,macina-etal-2023-mathdial}, and scientific concept education~\cite{yuan-etal-2024-boosting,yang2024leveraging}. 
Most aim to enhance students' understanding of target knowledge by employing pedagogical strategies such as recommending exercises~\cite{deng2023towards} or selecting teaching examples~\cite{ross-andreas-2024-toward}. 
However, these approaches fall short in broader \textit{task tutoring} situations demanding both understanding and practical application of specific pieces of knowledge to solve real-world, goal-driven problems. 
Such scenarios require tutors to proactively guide people toward completing targeted tasks (e.g., coding).
Furthermore, the tutoring outcomes are challenging to assess since targeted tasks can often be completed with open-ended solutions.

To bridge this gap, we introduce \textbf{coding tutoring}, a promising yet underexplored task for LLM agents.
As illustrated in Figure~\ref{fig:example}, the tutor is provided with a target coding task and task-specific knowledge (e.g., cross-file dependencies and reference solutions), while the student is given only the coding task. The tutor does not know the student's prior knowledge about the task.
Coding tutoring requires the tutor to proactively guide the student toward completing the target task through dialogue.
This is inherently a goal-oriented process where tutors guide students using task-specific knowledge to achieve predefined objectives. 
Effective tutoring requires personalization, as tutors must adapt their guidance and communication style to students with varying levels of prior knowledge.

Developing effective tutoring agents is challenging because off-the-shelf LLMs lack grounding to task-specific knowledge and interaction context.
Specifically, tutoring requires \textit{epistemic grounding}~\citep{tsai2016concept}, where domain expertise and assessment can vary significantly, and \textit{communicative grounding}~\citep{chai2018language}, necessary for proactively adapting communications to students' current knowledge.
To address these challenges, we propose the \textbf{Tra}ce-and-\textbf{Ver}ify (\textbf{\model}) agent workflow for building effective LLM-powered coding tutors. 
Leveraging knowledge tracing (KT)~\citep{corbett1994knowledge,scarlatos2024exploring}, \model explicitly estimates a student's knowledge state at each turn, which drives the tutor agents to adapt their language to fill the gaps in task-specific knowledge during utterance generation. 
Drawing inspiration from value-guided search mechanisms~\citep{lightman2023let,wang2024math,zhang2024rest}, \model incorporates a turn-by-turn reward model as a verifier to rank candidate utterances. 
By sampling more candidate tutor utterances during inference (see Figure~\ref{fig:scale}), \model ensures the selection of optimal utterances that prioritize goal-driven guidance and advance the tutoring progression effectively. 
Furthermore, we present \textbf{Di}alogue for \textbf{C}oding \textbf{T}utoring (\textbf{\eval}), an automatic protocol designed to assess the performance of tutoring agents. 
\eval employs code generation tests and simulated students with varying levels of programming expertise for evaluation. While human evaluation remains the gold standard for assessing tutoring agents, its reliance on time-intensive and costly processes often hinders rapid iteration during development. 
By leveraging simulated students, \eval serves as an efficient and scalable proxy, enabling reproducible assessments and accelerated agent improvement prior to final human validation.

Through extensive experiments, we show that agents developed by \model consistently demonstrate higher success rates in guiding students to complete target coding tasks compared to baseline methods. We present detailed ablation studies, human evaluations, and an inference time scaling analysis, highlighting the transferability and scalability of our tutoring agent workflow.

\section{Problem Definition}

\begin{figure*}[th!]
\centering
\includegraphics[width=1.0\textwidth]{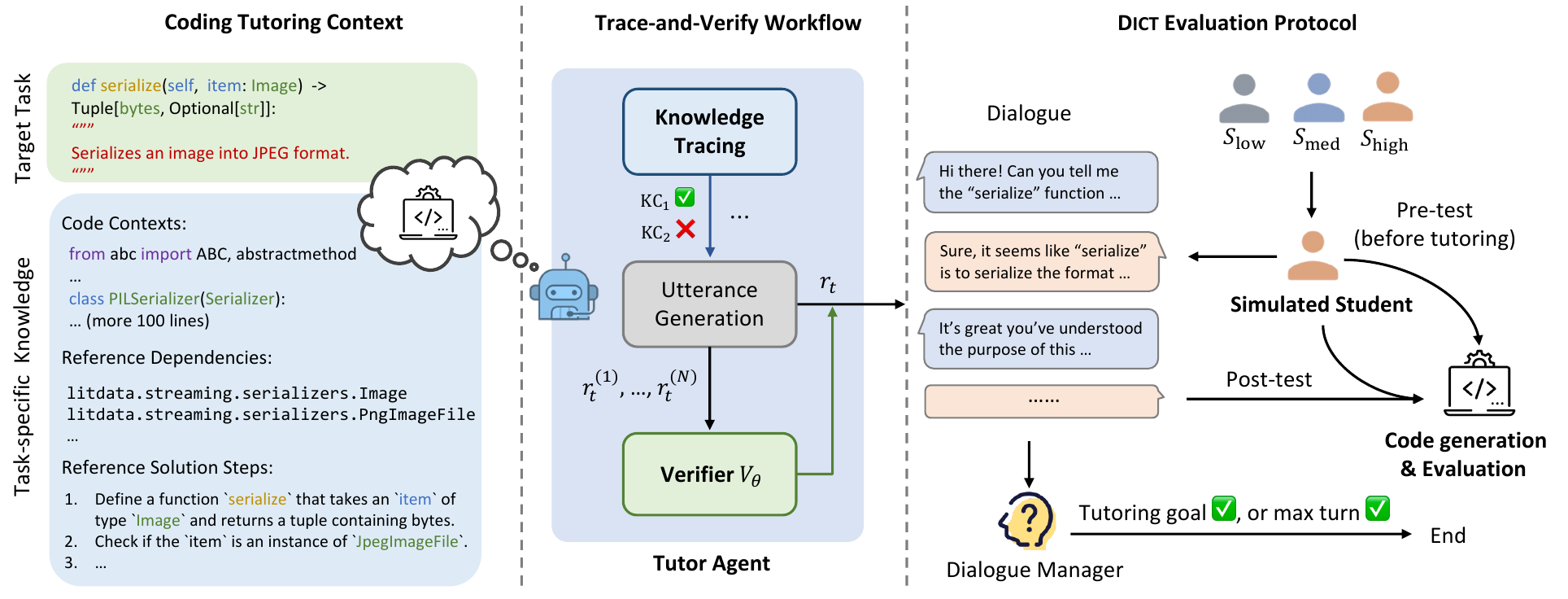}
\caption{Overview of our work for developing coding tutoring agents. \textit{Left}: The context of the coding tutoring problem. \textit{Middle}: Trace-and-Verify (\textsc{Traver}) workflow. \textit{Right}: \textsc{Dict} evaluation protocol.
}
\label{fig:overview}
\end{figure*}

We formulate \textbf{coding tutoring} as an interactive dialogue process between a \textbf{tutor} and a \textbf{student}, where the goal is to help the student implement a working solution that passes predefined unit tests for a target coding task.

Formally, the tutor is assigned a coding task $\mathcal{T}$ that consists of a function signature and a requirement description outlining the desired functionality. 
The tasks are repository-level, which require an understanding of multiple interdependent files within the codebase to implement a correct solution.
The tutor has access to task-specific knowledge $\mathcal{K}$, which includes (\romannumeral1) \textit{Code Contexts}: Contextual code snippets surrounding the desired code, which help the tutor show examples when necessary; (\romannumeral2) \textit{Reference Dependencies}: Cross-referenced elements such as intra-class, intra-file, and cross-file dependencies, along with their corresponding descriptions (e.g., docstrings), which involve key knowledge for completing the desired code; and (\romannumeral3) \textit{Reference Solution Steps}: Key steps required to complete the target task, describing using natural languages. 

The student is given the task $\mathcal{T}$ and possesses some subset of $\mathcal{K}$ as their prior knowledge, but the tutor remains unaware of which specific concepts or dependencies the student has already mastered.
The goal of the tutor is to guide the student, regardless of his or her background, toward solving the task $\mathcal{T}$ through multi-turn interactions.
\section{\textbf{Tra}ce-and-\textbf{Ver}ify Agent Workflow}
\label{sec:methods}

We propose \textbf{Tra}ce-and-\textbf{Ver}ify (\textbf{\model}), an effective workflow for developing tutor agents (see the middle part in Figure \ref{fig:overview}). 
\model integrates two key components: (\romannumeral1) explicit tracing of a student's knowledge state and (\romannumeral2) utterance decoding guided by a verifier model for turn-by-turn verification.

\subsection{Adapting to Student's Knowledge via Knowledge Tracing}
Effective tutoring requires bridging the gap between a student's prior knowledge and the skills needed to solve the target coding task. 
To address this, we employ knowledge tracing (KT)~\citep{corbett1994knowledge,abdelrahman2023knowledge,scarlatos2024exploring} to estimate the student's knowledge state at each dialogue turn.
Specifically, we represent task-specific knowledge $\mathcal{K}$ as a set of knowledge components (KCs) $\{\text{KC}_1, \text{KC}_2, \dots, \text{KC}_K\}$, where each KC is either a reference dependency or a solution step. 
At the $t$-th turn, the tutor agent explicitly assesses the student's belief of each KC using texts, based on the dialogue context $\mathcal{C}_t$ and the estimated belief $B_{t-1}$ at the previous turn. 
The current belief $B_{t}$ indicates how many KCs the student has understood. 
With this estimation, the tutor is further prompted to focus more on missing KCs and generate utterances that address the student's knowledge gaps. 
The detailed prompt template is provided in Appendix~\ref{appendix:prompt_KT}.

\subsection{Utterance Decoding via Turn-by-Turn Verification}
Based on the KT outcomes, the tutor agent aims to generate high-quality utterances that advance the tutoring process. 
However, LLMs often struggle to determine which utterances effectively guide students toward task completion. 
Drawing inspiration from value-guided search approaches~\citep{lightman2023let,wang2024math,zhang2024rest}, we address this with a turn-by-turn verifier.
The verifier $V_{\theta}$ evaluates the quality of potential tutor responses by producing a reward score $v_{t}\in[0,1]$ based on three inputs: the target task $\mathcal{T}$, current dialogue context $\mathcal{C}_t$, and a candidate tutor utterance $r_t$ at turn $t$. To select the optimal response, we generate $N$ candidate utterances through parallel sampling and choose the one that receives the highest reward score from the verifier.

The core of the verifier $V_{\theta}$ is the turn-based reward $v_t$ at $t$-th turn, which should reflect (\romannumeral1) the cumulative progress made in the previous turns and (\romannumeral2) the current turn's contribution to achieving the overall tutoring goal. Hence, $v_{t}$ can be iteratively defined as:
\begin{equation}
    v_t = \max(v_{t-1} + w_{r_t}, 0) ~(v_0=0, t\in[1,T])
\end{equation}
where $w_{r_t}$ is the weighted reward quantifying the contribution of the current turn to the overall goal, $T$ denotes the total number of turns. 
To compute $w_{r_t}$, we introduce the concept of guiding distance $d_t=T-t$, which measures the remaining turns until the goal or end of the interaction. The weighted reward is then calculated as:
\begin{equation}
\label{eq:weighted_reward}
    w_{r_t} = \frac{1-v_{t-1}}{d_{t}+1}(2o_{s_t}-1)
\end{equation}
where $o_{s_t}\in\{0,1\}$ is a binary outcome indicating whether the tutor's $t$-th utterance contributes to the student's eventual successful completion of the target task.
As the dialogue progresses, the guiding distance $d_t$ decreases, leading to larger weighted rewards for later turns. This design ensures the turn-based reward $v_t$ remains bounded while appropriately weighting the importance of each turn based on its proximity to the goal.

We train the verifier $V_{\theta}$ using mean squared error (MSE) loss over $n$ samples:
\begin{equation}
    \mathcal{L} = \frac{1}{n}\sum_{i=1}^{n}\sum_{t=1}^{T_i}\left(V_{\theta}([\mathcal{T}^{(i)};\mathcal{C}_{t}^{(i)};r_{t}^{(i)}]) - v_{t}^{(i)}\right)^2
\end{equation}
where $T_{i}$ denotes the number of turns for the $i$-th sample, $\theta$ denotes the trainable parameters.

During inference, the verifier serves as a plug-and-play module, which chooses the utterance with the highest reward from candidate utterances generated by parallel sampling at each turn, promoting progression toward tutoring task completion.

\section{The \eval Evaluation Protocol}
\label{sec:evaluation}

One key challenge in developing tutoring agents is the lack of robust evaluation methods. 
While human evaluation is essential, its high cost, time requirements, and complexity make it impractical for scalable benchmarking.
To address this limitation, we present \textbf{Di}alogue for \textbf{C}oding \textbf{T}utoring (\textbf{\eval}), an automatic protocol for evaluating tutor agents. The overview of \textsc{Dict} is shown in Figure \ref{fig:overview}.
Our protocol employs LLMs to simulate students with varying levels of programming knowledge. First, tutors engage in multi-turn dialogues with students to tutor the task. Then, students demonstrate their learning outcome by implementing the solution code. We evaluate tutoring effectiveness through automated unit tests of the student-generated code. This automated approach enables controlled, reproducible, and scalable evaluations of tutoring agents.

\subsection{Controlled Student Simulation}
\label{sec:student_simulation}
To evaluate how well tutors can adapt their strategies to students with varied prior knowledge, we simulate students of three knowledge levels:
(1) \textbf{Low-level}: Students access no prior knowledge from $\mathcal{K}$. They represent beginners with no familiarity with the target task. (2) \textbf{Medium-level}: These students are assigned a proportion (e.g., 50\%) of the \textit{reference dependencies} by random sampling, denoting that they have partial knowledge required for completing the task. (3) \textbf{High-level}: In addition to partial \textit{reference dependencies}, these students are also provided with the \textit{code contexts}, indicating that they have more comprehensive knowledge with contextual guidance.

\paragraph{Pre-Test.}
\label{sec:pre_test} 
We create students of all three levels using the same LLM simulator, varying only in their knowledge level. 
However, this raises a critical question: \textit{do students with different knowledge levels actually demonstrate distinct performance in completing the coding tasks?} To validate this, we conduct a preliminary coding test where each simulated student attempts to generate code for the target task $\mathcal{T}$ before any tutoring intervention (see Appendix \ref{appendix:prompt} for prompting details).

\paragraph{Metrics for Coding Test.}
\label{sec:metrics_coding}
Following previous studies \cite{austin2021program,li2024evocodebench}, we employ Recall@$k$ and Pass@$k$ as evaluation metrics to assess coding test performance. Recall@$k$ measures the recall of reference dependencies in the generated programs. Specifically, students are asked to generate $k$ programs per target task. For the $i$-th program, we extract its dependencies $\mathbb{P}_{i}$ using the Pyan parser \cite{pyan2023}. We compare them with the reference dependencies $\mathbb{R}$ and compute the Recall@$k$ as:
\begin{equation}
\label{eq:recall}
\text{Recall}@k = \underset{\text{Target Tasks}}{\mathbb{E}} \left[ \max_{i \in [1, k]} \frac{|\mathbb{R} \cap \mathbb{P}_i|}{|\mathbb{R}|}\right]
\end{equation}
where $|\cdot|$ denotes the number of elements of a set. 
Pass@$k$ evaluates the functional correctness of the generated programs.
After generating $n \geq k$ programs per task, we execute them in Python interpreters to count the number of correct programs $c$ that pass all test cases. Pass@$k$ is computed as:
\begin{equation}
\label{eq:pass}
\text{Pass}@k =\underset{\text{ Target Tasks}}{\mathbb{E}}\left[1-\frac{\left(\begin{array}{c}
n-c \\
k
\end{array}\right)}{\left(\begin{array}{l}
n \\
k
\end{array}\right)}\right]
\end{equation}
where $\binom{\cdot}{\cdot}$ denotes the number of ways to choose a subset of elements (also known as the binomial coefficient). Our pre-test results reported in \S\ref{sec:student_analysis} show that the simulated students are effective.

\subsection{Tutor-Student Interaction}
\label{sec:tutor_student}
As shown in Figure \ref{fig:overview}, we let an LLM-based tutor agent engage in a multi-turn dialogue with a chosen student, simulating a tutoring session. The tutor is initialized with the target coding task $\mathcal{T}$ and task-specific knowledge $\mathcal{K}$. We ask the tutor to initiate the tutoring and communicate with the student turn by turn. A key challenge is determining when to terminate the tutoring. While the tutor could self-determine, our preliminary experiments revealed a tendency for overconfidence, leading to premature termination. This issue arises because most tutors overlook gaps in the student's prior knowledge. For a robust comparison, we follow \citet{wang-etal-2023-target} and introduce an LLM-powered dialogue manager (see Figure \ref{fig:overview}). Operating from a ``God's perspective,'' the manager considers the dialogue context and all information from both the tutor and student, to decide whether the tutoring goal has been met. The tutoring terminates under one of two conditions: (\romannumeral1) The manager confirms that the tutoring goal is achieved; (\romannumeral2) The dialogue reaches a predefined maximum of $T$ turns.

\subsection{Automatic Evaluation}

\paragraph{Post-Test.}
To evaluate the effectiveness of tutor agents, we conducted a coding test after tutoring (referred to as the post-test). Given a target task $\mathcal{T}$ and a dialogue session $\mathcal{C}=\{s_t, r_t\}_{t=1}^{T}$ with a simulated student, we ask the student to generate code for fulfilling the task. 
However, assuming that all dialogue content is retained during the coding test may be unrealistic. According to cognitive load theory (CLT) \cite{miller1956magical,sweller2011cognitive}, human working memory has a limited learning capacity at one time, and exceeding this capacity can hinder learning. This cognitive load can be affected by the complexity of the target task or student engagement during the tutoring dialogue.

As a simple, practical alternative, we consider the student's cognitive load $f_{\text{CL}}$ at each turn by assuming the information retained from the tutor’s utterance to a maximum threshold. Specifically, if the tutor's utterance $r_t$ exceeds $M$ words, only the latest $M$ words are retained; otherwise, the full utterance is kept. Our post-test is formulated as:
\begin{equation}
\label{eq:post_test}
    \mathcal{Y}_{\text{code}}=\text{LM}_{\text{Student}}([\mathcal{I};\mathcal{T};\{s_t, f_{\text{CL}}(r_t)\}_{t=1}^{T}])
\end{equation}
where $\mathcal{I}$ represents the instruction for code generation. $s_t$ and $r_t$ denote student and tutor utterances at $t$-th turn, respectively. A detailed template for the instruction can be found in Appendix \ref{appendix:prompt}.

\begin{table*}[t!]
\centering
\scalebox{0.85}{
\begingroup
\renewcommand{\arraystretch}{0.85}
\setlength{\tabcolsep}{4pt}
\hspace{-10pt}
\begin{tabular}{cl  cc  cc cc cc}
\toprule
\multirow{2}{*}{\textbf{Method}} & \multirow{2}{*}{\textbf{Backbone Model}} & \multicolumn{2}{c}{\textbf{Overall}} &  \multicolumn{2}{c}{\textbf{Low-level}} & \multicolumn{2}{c}{\textbf{Med.-level}}  & \multicolumn{2}{c}{\textbf{High-level}}   \\
 \cmidrule(lr){3-4} \cmidrule(lr){5-6} \cmidrule(lr){7-8} \cmidrule(lr){9-10}
 &   &   Recall / $\Delta\%$R & Pass / $\Delta\%$P & $\Delta\%$R & $\Delta\%$P  &  $\Delta\%$R & $\Delta\%$P &  $\Delta\%$R & $\Delta\%$P  \\
\cmidrule(lr){1-2}\cmidrule(lr){3-4} \cmidrule(lr){5-6} \cmidrule(lr){7-8} \cmidrule(lr){9-10}
\rowcolor{gray!10}
Pre-Test & --  & 45.9\textsubscript{$\pm~2.8$} / -- & 21.2\textsubscript{$\pm~2.0$} / -- &  --  & --  & --  & --  &  --  & --  \\
\cmidrule(lr){1-2}\cmidrule(lr){3-4} \cmidrule(lr){5-6} \cmidrule(lr){7-8} \cmidrule(lr){9-10}
\multirow{7}{*}{\shortstack{Vanilla \\ Instruct}}  & Qwen2-7B-Instruct & 56.4\textsubscript{$\pm~1.6$} / 22.8 & 26.4\textsubscript{$\pm~3.4$} / 24.5 &  99.4  & 69.8  & 10.1  & 37.5 & 3.0 & -0.3  \\
  & Qwen2-72B-Instruct  & 61.4\textsubscript{$\pm~1.9$} / 33.8 & 32.1\textsubscript{$\pm~7.0$} / 51.4 & 131.8 & 128.2  & 13.7  & 53.7  &  11.5  & 21.4  \\
  & Llama-3.1-8B-Instruct & 63.6\textsubscript{$\pm~4.4$} / 38.5 & 31.1\textsubscript{$\pm~3.3$} / 46.7 & 138.6 &  145.1  & 23.8  & 48.3 & 10.9  & 8.0     \\
  & Llama-3.1-70B-Instruct & 62.5\textsubscript{$\pm~4.0$} / 36.0  & 34.9\textsubscript{$\pm~5.4$} / 65.0 & 127.4  & 160.6 & 21.5   & 71.0 & 11.7  & 24.9   \\
  & GPT-3.5-Turbo  & 60.1\textsubscript{$\pm~4.0$} / 31.0 & 28.8\textsubscript{$\pm~3.6$} / 35.9 & 110.9 & 130.1  & 21.6 & 37.7 & 6.9  & -1.3  \\
  & GPT-4o  & 64.2\textsubscript{$\pm~4.3$} / 39.9 & 38.7\textsubscript{$\pm~5.6$} / 82.8 & 141.3 & 207.8  & 28.4  & 102.1 & 8.9 & 23.5 \\
  & o1-mini & 61.3\textsubscript{$\pm~2.1$} / 33.4 & 35.9\textsubscript{$\pm~1.4$} / 69.4 & 129.5  & 159.1  & 19.3  & 101.2  & 6.9  & 16.6 \\
\cmidrule(lr){1-2}\cmidrule(lr){3-4} \cmidrule(lr){5-6} \cmidrule(lr){7-8} \cmidrule(lr){9-10}
Self-Refine & GPT-4o & 64.0\textsubscript{$\pm~4.9$} / 39.5  & 40.6\textsubscript{$\pm~3.7$} / 91.7  & 143.0 & 221.4  & 23.1 & 118.6 & 11.8 & 26.2 \\
\cmidrule(lr){1-2}\cmidrule(lr){3-4} \cmidrule(lr){5-6} \cmidrule(lr){7-8} \cmidrule(lr){9-10}
TreeInstruct & GPT-4o  & 64.3\textsubscript{$\pm~2.8$} / 40.1 &  39.8\textsubscript{$\pm~1.5$}  / 88.1 &  154.8  & 211.5 & 21.8  & 100.3  & 9.7  & 33.6  \\
\cmidrule(lr){1-2}\cmidrule(lr){3-4} \cmidrule(lr){5-6} \cmidrule(lr){7-8} \cmidrule(lr){9-10}
\multirow{2}{*}{\shortstack{\textbf{\model} \\ \textbf{(Ours)}}} & Llama-3.1-70B-Instruct &  66.8\textsubscript{$\pm~1.3$} / 45.5 & 39.3\textsubscript{$\pm~6.9$} / 85.7  &  164.5  & 206.9  & 23.4  & 104.1  & \textbf{16.4}  & 28.4      \\
 &  GPT-4o  & \textbf{68.8}\textsubscript{$\pm~3.7$} / \textbf{49.8} & \textbf{43.7}\textsubscript{$\pm~1.3$} / \textbf{106.5} &  \textbf{166.3}  & \textbf{242.5}  & \textbf{34.8}  & \textbf{122.9}  & 15.8  & \textbf{44.8}  \\
\cmidrule(lr){1-2}\cmidrule(lr){3-4} \cmidrule(lr){5-6} \cmidrule(lr){7-8} \cmidrule(lr){9-10}
\rowcolor{gray!10}
Oracle & --  & 74.0\textsubscript{$\pm~5.7$} / 61.2  &  51.9\textsubscript{$\pm~3.7$} / 144.8 & 200.8 & 318.5  & 42.3 & 176.1 & 21.1  & 60.7 \\
\bottomrule
\end{tabular}
\endgroup}
\caption{Automatic evaluation results of various LLM-based tutor agents. ``$\Delta\%$R'' and ``$\Delta\%$P'' represent the tutoring outcome rates (TOR) in Recall and Pass, respectively.}
\label{tab:overall_result}
\end{table*}

\paragraph{Evaluation Metrics.}
Based on the coding test, students' post-test performance after tutoring is defined as the \textbf{tutoring outcome} (\textbf{TO}), measured by \textbf{Recall} and \textbf{Pass}. They represent the averages of Recall@$k$ and Pass@$k$ for $k\in\{1,3,5,10\}$. A higher Recall score indicates the tutor is more capable of leading the student to acquire the dependency knowledge for coding; a higher Pass score denotes a higher success rate of guiding the student in completing the target coding task.

Due to the difference in prior knowledge levels, we use the \textbf{tutoring outcome rate} (\textbf{$\Delta\%$}) to normalize and fairly evaluate the tutor's performance. This is calculated by relative improvement before and after tutoring:
\begin{equation}
    \Delta\%\mathcal{M}= \left(\frac{\mathcal{M}_{\text{post-test}}-\mathcal{M}_{\text{pre-test}}}{\mathcal{M}_{\text{pre-test}}}\right) \times 100\%
\end{equation}
where $\mathcal{M}$ denotes the metric for the coding test, which can be either Recall or Pass.

To further analyze the tutoring process, we propose the \textbf{tutoring outcome curve} (\textbf{TOC}). At each $t$-th turn, we ask simulated students to perform a post-test using the dialogue context up to that turn, i.e., $\{s_{\leq t},r_{\leq t}\}$. The TOC is then plotted by tracking the Recall and Pass scores varying by turns. These curves exhibit how tutor agents guide students throughout the tutoring session.

\section{Experiments}

\subsection{Experimental Setup}

\begin{figure*}[t!]
\centering
\includegraphics[width=0.99\textwidth]{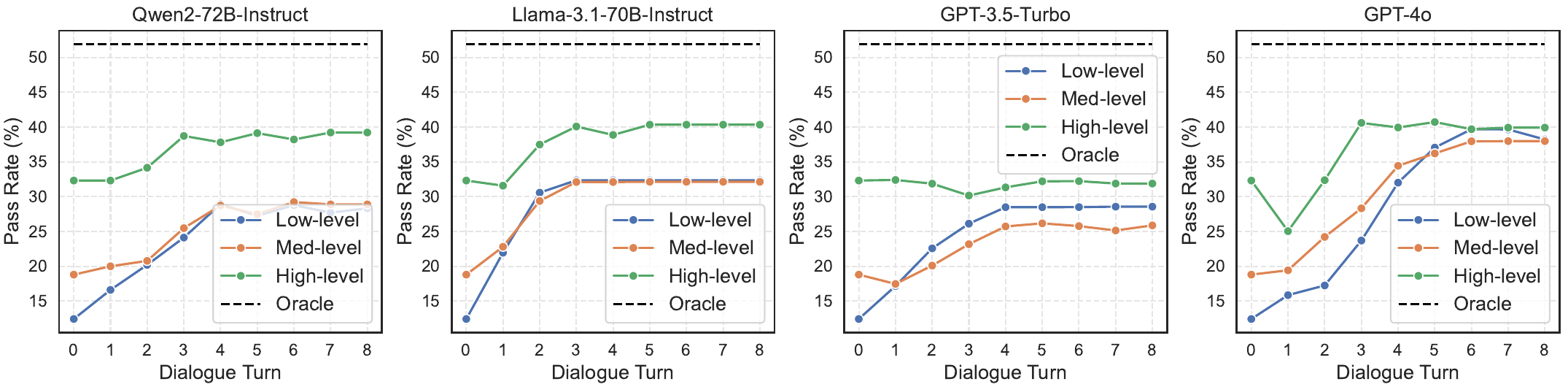}
\caption{Tutoring outcome curves in Pass rate across various LLM-based tutors with Vanilla Instruct.}
\label{fig:toc_base}
\vspace{-4pt}
\end{figure*}

\paragraph{Benchmark.}
We adopt \textbf{EvoCodeBench} \cite{li2024evocodebench} as the testbed due to its realistic repository-level Python coding tasks along with dependency annotations and repository contexts, providing a rich knowledge foundation for coding tutoring. 
We have 100 target coding tasks and split them equally into 5 folds, using 5-fold cross-validation for experiments. The detailed examples, statistics, and preprocessing are provided in Appendix \ref{appendix:dataset}.

\paragraph{Student Simulator.}
Prior to tutoring, it is essential to ensure that simulated students have not been exposed to any target coding task. To avoid data contamination, we use Mixtral-8x7B-Instruct \cite{jiang2024mixtral} as the student simulator. This model's training data only includes content up to \texttt{2023-9}, whereas all coding tasks in EvoCodeBench are collected from the repositories created between \texttt{2023-10} and \texttt{2024-2}. Furthermore, this model has strong conversational and coding abilities, making it well-suited for student simulation.
Since different simulators can influence tutoring outcomes, we also use GPT-4o~\cite{openai2024gpt4o} as the student simulator and report experimental results in Appendix~\ref{appendix:add_result}.

\paragraph{Backbone Models.}
We employ various backbone models to develop tutor agents, including: \textbf{Qwen2-Instruct} \cite{yang2024qwen2} with 7B and 72B variants, \textbf{Llama-3.1-Instruct} \cite{dubey2024llama} with 8B and 70B variants, \textbf{GPT-3.5-Turbo} \cite{openai2022chatgpt}, \textbf{GPT-4o} \cite{openai2024gpt4o}, and \textbf{o1-mini} \cite{openai2024o1}.

\paragraph{Baseline Methods}
We evaluate our \model, against the following relevant baseline methods: (1) \textbf{Vanilla Instruct}: It directly instructs LLMs as tutors, as detailed in \S\ref{sec:tutor_student}. (2) \textbf{Self-Refine}~\cite{madaan2023self}: LLMs generate initial responses and iteratively refine these responses by providing natural language feedback to themselves. (3) \textbf{TreeInstruct}~\cite{kargupta-etal-2024-instruct}: It is a Socratic teaching method that estimates a student's knowledge and employs tree-based questioning to guide the student. (4) \textbf{Oracle}: It provides the full task-specific knowledge directly to each student during the post-test (see Eq. (\ref{eq:post_test})).

\paragraph{Implementation Details.}
We implement the verifier in \model using Mistral-7B \cite{jiang2023mistral} with an additional linear layer. We utilize the synthesized dialogues from various backbone LLMs with vanilla instructions for the verifier, where the post-test results on Pass scores provide outcome reward labels. We adopt 5-fold cross-validation for training and evaluation. 
In the \eval evaluation protocol, the maximum number of turns $T$ is set to 8. The cognitive load parameter $M$ is set to 60 during the post-test.
More details about training and inference are provided in Appendix \ref{appendix:implementation}.

\subsection{Experimental Results}
\label{sec:exp_results}

\paragraph{How do various LLMs perform as tutor agents when provided with vanilla instructions?}
As shown in Table~\ref{tab:overall_result}, various LLM-based tutors with vanilla instructions perform significantly inferior to the Oracle tutor, indicating clear limitations. Scaling the parameter size of open-source models like Qwen2 and Llama-3.1 generally improves performance. However, the large gaps in Pass and TOR-Pass scores suggest that simply using larger models is inadequate to guide students in successfully completing target coding tasks. 
These findings indicate that developing effective tutor agents requires not only detailed instructions but also mechanisms to facilitate tutoring outcomes in a structured way.

Another limitation that emerges from these results is the adaptability. 
As shown in Table~\ref{tab:overall_result}, LLMs generally perform better with low-level than with high-level students. This discrepancy arises from the greater difficulty in adapting to higher-level students who require nuanced and targeted guidance. For example, Qwen2-7B-Instruct and GPT-3.5-Turbo show a decline in Pass rates for high-level students after tutoring (i.e., $\Delta\%\text{P}<0$). Figure~\ref{fig:toc_base} illustrates this trend by tutoring outcome curves. Larger models like Qwen2-72B-Instruct and Llama3-3.1-70B-Instruct exhibit significant performance gaps between high-level students and others, even as tutoring dialogues progress. While GPT-4o demonstrates better adaptability, its tutoring outcomes early plateau from the third turn for high-level students, indicating diminishing returns. These results highlight the importance of enhancing tutor agents to adaptively guide different students.

\begin{figure}[t!]
\centering
\includegraphics[width=1\linewidth]{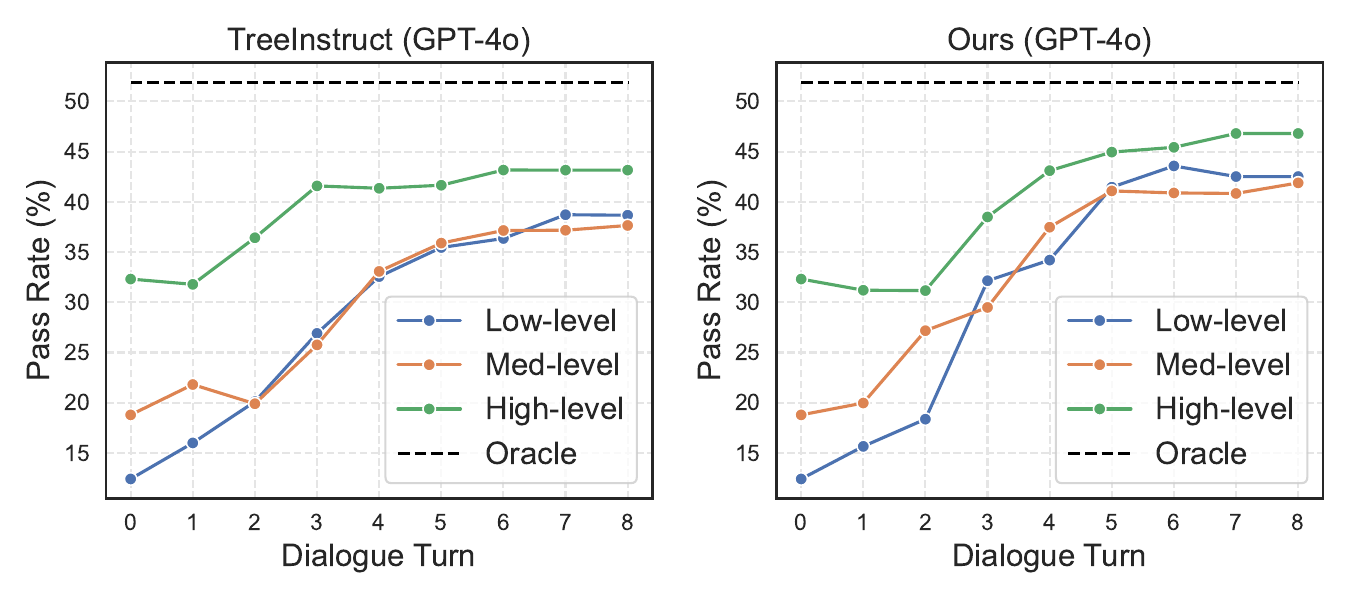}
\caption{Comparison of tutoring outcome curves between the TreeInstruct and \model (Ours).}
\label{fig:toc_comp}
\end{figure}

\begin{table}[t!]
\centering
\scalebox{0.82}{
\begingroup
\renewcommand{\arraystretch}{0.9}
\setlength{\tabcolsep}{3pt}
\hspace{-8pt}
\begin{tabular}{l  cc}
\toprule
\textbf{Backbone Model} &  \textbf{Recall} / \textbf{$\Delta\%$R} & \textbf{Pass} / \textbf{$\Delta\%$P}  \\
\cmidrule(lr){1-1}\cmidrule(lr){2-3}
Llama-3.1-70B-Instruct &  \textbf{66.8}\textsubscript{$\pm~1.3$} / \textbf{45.5}  & \textbf{39.3}\textsubscript{$\pm~6.9$} / \textbf{85.7}  \\
~~ w/o KT & 66.1\textsubscript{$\pm~3.4$} / 43.9  & 35.8\textsubscript{$\pm~3.8$} / 69.2  \\
~~ w/o Verifier  & 66.7\textsubscript{$\pm~3.2$} / 45.3  &  35.1\textsubscript{$\pm~5.1$} / 65.8  \\
\cmidrule(lr){1-1}\cmidrule(lr){2-3}
GPT-4o & \textbf{68.8}\textsubscript{$\pm~3.7$} / \textbf{49.8} &  \textbf{43.7}\textsubscript{$\pm~1.3$} / \textbf{106.5}  \\
~~ w/o KT & 65.9\textsubscript{$\pm~2.2$} / 43.5  & 41.7\textsubscript{$\pm~2.7$} / 97.1   \\
~~ w/o Verifier   &  67.8\textsubscript{$\pm~0.8$} / 47.7  & 39.8\textsubscript{$\pm~3.2$} / 88.2  \\
\bottomrule
\end{tabular}
\endgroup}
\caption{Ablation study results of our \model.}
\label{tab:ablation_result}
\vspace{-4pt}
\end{table}

\paragraph{Can \model improve tutoring outcomes and better adapt to different students?}
As shown in Table~\ref{tab:overall_result}, our \model, built upon Llama-3.1-70B-Instruct, achieves notable improvements over Vanilla Instruct (e.g., from 34.9\% to 39.3\% in Pass rate). When compared to tutor agents built upon GPT-4o, \model achieves the highest overall Recall and Pass rates. 
These results highlight that our approach is more effective in guiding students to successfully complete target coding tasks. More importantly, \model exhibits substantial improvements across students with different levels, narrowing the gap with the Oracle tutor. The tutoring outcome curves in Figure~\ref{fig:toc_comp} further illustrate that, regardless of students' prior knowledge levels, our method consistently improves the success rate of guiding students to achieve task completion.

\begin{table*}[t!]
\centering
\scalebox{0.82}{
\begingroup
\renewcommand{\arraystretch}{0.8}
\setlength{\tabcolsep}{4pt}
\hspace{-8pt}
\begin{tabular}{l  ccc  ccc  ccc}
\toprule
\multirow{2}{*}{\textbf{Compared Methods}} & \multicolumn{3}{c}{\textbf{Proactivity}} &  \multicolumn{3}{c}{\textbf{Adaptability}} & \multicolumn{3}{c}{\textbf{Coherence}}   \\
\cmidrule(lr){2-4} \cmidrule(lr){5-7}\cmidrule(lr){8-10}  
  &  Win (\%) & Lose (\%) & Tie (\%) &  Win (\%) & Lose (\%) & Tie (\%) &  Win (\%) & Lose (\%) & Tie (\%) \\
\cmidrule(lr){1-1}\cmidrule(lr){2-4} \cmidrule(lr){5-7}\cmidrule(lr){8-10}  
\model vs. Vanilla Instruct  &  \textbf{42.2} & 26.7 & 31.1  & \textbf{40.0} & 33.3 &  26.7  & 24.4  & \textbf{25.6}  &  50.0 \\
\model vs. Self-Refine  & \textbf{38.9} & 26.7  &  34.4  & \textbf{41.1}  &  30.0  &  28.9  & \textbf{27.8}  &  18.9  &  53.3  \\
\model vs. TreeInstruct   & \textbf{34.4} & 22.2  &  43.3  &  \textbf{37.8}  & 25.6  &  36.7  &  \textbf{28.9}  &  20.0  & 51.1   \\
\bottomrule
\end{tabular}
\endgroup}
\caption{Human evaluation results. For win and lose percentages, the higher value is bolded.}
\label{tab:human_eval_result}
\vspace{-4pt}
\end{table*}

\paragraph{Ablation study.}
We conduct an ablation study for \model: (1) \textit{\textbf{w/o knowledge tracing (KT)}}, which removes the KT operation prior to tutor utterance generation; and (2) \textit{\textbf{w/o verifier}}, which omits the verifier $V_{\theta}$ used for verification. 
The results in Table~\ref{tab:ablation_result} show that both the KT and verifier contribute to the overall performance. 
Notably, the absence of the verifier leads to a sharp decline in Pass and TOR-Pass, underscoring its critical role. 
The verifier improves the possibility of generated utterances that effectively advance the tutoring progress at each turn, thereby increasing the success rate in guiding students to complete the coding task.

\begin{figure}[t!]
\centering
\includegraphics[width=1\linewidth]{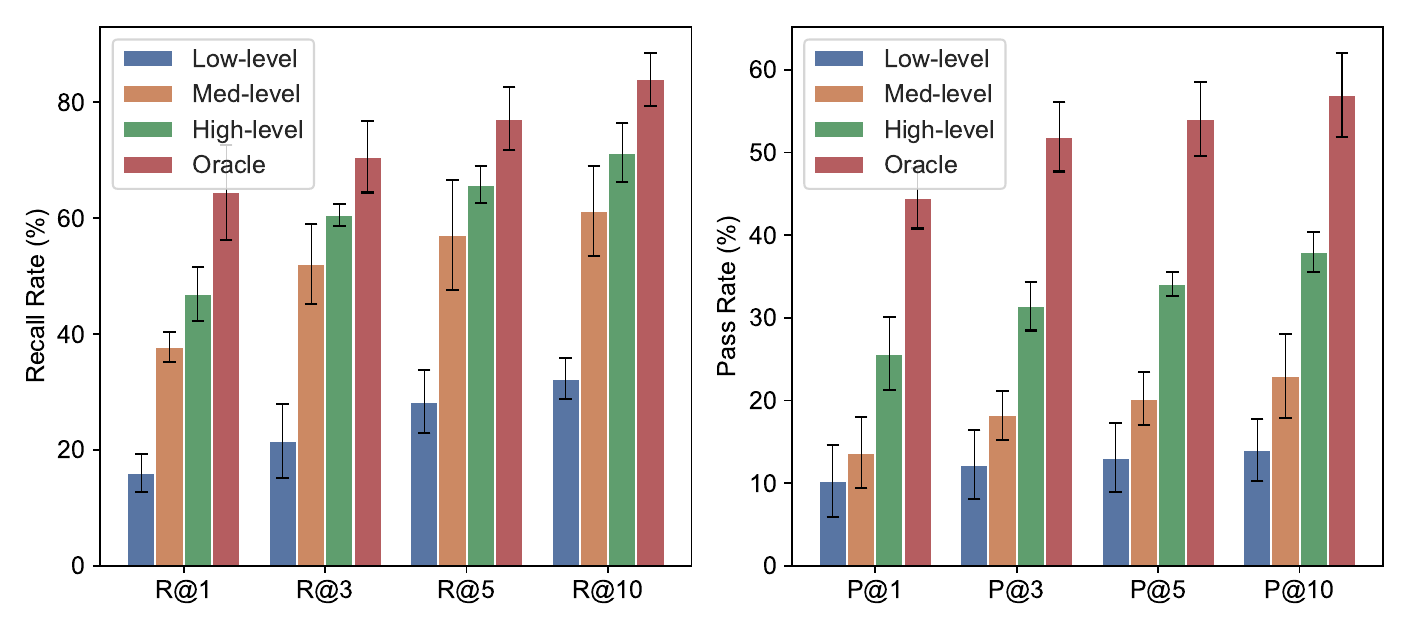}
\caption{Pre-test performance of simulated students at different levels before tutoring.}
\label{fig:pretest_result}
\vspace{-4pt}
\end{figure}

\subsection{Analysis of Simulated Students}
\label{sec:student_analysis}

Since the \eval evaluation protocol relies on the LLM-simulated students, it is crucial to ensure that these students, with varying levels of prior knowledge, align with discrepancies in completing coding tasks. To validate this, we conducted a pre-test for the simulated students before tutoring, as described in~\S\ref{sec:pre_test}. Figure \ref{fig:pretest_result} shows that simulated low-, medium-, and high-level students exhibit significant performance differences in terms of Recall@$k$ and Pass@$k$. Under a controlled setup, students at different levels demonstrate distinct abilities in completing target coding tasks. Therefore, our student simulation serves as a feasible proxy for real students, offering its advantages of scalability and cost-effectiveness for evaluating tutor agents.

\subsection{Inference-Time Scaling with Verifier}
\label{sec:scaling_analysis}
Using Llama-3.1-70B-Instruct as the backbone model, we vary the number of candidate tutor utterances per turn, i.e., $N$, within $\{1, 5, 10, 15, 20\}$ and ask the verifier to select the best one based on predicted rewards. 
We also evaluate two baselines: (\romannumeral1) Chain-of-Thought (CoT)~\citep{wei2022chain} prompting, which performs zero-shot CoT reasoning (i.e., ``Let's think step by step and select the best response.'') to choose the most appropriate tutor utterance from a set of candidates at each turn (denoted as B1), and (\romannumeral2) random selection (denoted as B2).
As shown in Figure~\ref{fig:scale}, \model with the trained verifier improves the student's Pass rate from 35.1\% to 39.3\% as $N$ increases, outperforming both the random and CoT baselines while exhibiting stronger linear scaling. 
Additionally, in terms of total tokens (including input prompts and output responses) consumed per tutoring session, \model achieves a better balance between tutoring performance and efficiency. These findings demonstrate that our \model effectively enables inference-time scaling for tutoring agents.

\subsection{Human Evaluation and Case Study}

To evaluate the quality of tutor agents developed by different methods, we conducted a human evaluation for Vanilla Instruct, Self-Refine, TreeInstruct, and our \model. 
We presented human evaluators with a pair of tutoring dialogues produced by two agents interacting with the same student. Evaluators were asked to determine which one is better from \textit{proactivity}, \textit{adaptability}, and \textit{coherence}. Further details are provided in Appendix~\ref{appendix:human_eval}.

Table~\ref{tab:human_eval_result} presents the evaluation results, with an average Fleiss’s kappa ($\kappa$) of 0.45, indicating moderate agreement among evaluators (0.41 < $\kappa$ < 0.60). The results demonstrate that our \model significantly outperforms the compared methods in proactivity and adaptability, while also matching or surpassing Vanilla Instruct in coherence. 
We provide several examples in Appendix~\ref{appendix:case}.
To further illustrate the performance of our tutor agents, we provide several examples in Appendix~\ref{appendix:case}.

\section{Related Work}

\paragraph{Interactive Tutoring.}
As an advanced form of tutoring systems, interactive intelligent tutoring systems (ITSs)~\citep{graesser2001intelligent,rus2013recent,liu-etal-2024-personality} can provide personalized feedback and adaptive learning experiences. 
They have been extensively explored across various educational domains for knowledge delivery, such as language learning~\citep{swartz2012intelligent,stasaski-etal-2020-cima,caines-etal-2020-teacher, kwon-etal-2024-biped}, mathematical reasoning~\citep{demszky-hill-2023-ncte,macina-etal-2023-mathdial,wang-etal-2024-bridging,liu2024socraticlm}, and scientific concept education~\citep{yuan-etal-2024-boosting,yang2024leveraging}. 
These studies mainly focus on enhancing students' understanding of specific pieces of knowledge, using pedagogical strategies such as designing exercises~\citep{deng2023towards,wang-etal-2022-towards,lu2023readingquizmaker}, selecting teaching examples~\citep{ross-andreas-2024-toward}, and remediating student reasoning errors~\citep{daheim-etal-2024-stepwise}.
Furthermore, the effectiveness of these approaches is often measured using closed-form assessments, such as question-answering~\cite{yuan-etal-2024-boosting} or multiple-choice tests~\cite{macina-etal-2023-mathdial}. 
Instead of focusing on specific knowledge delivery, we explore task-level tutoring, using coding tutoring as a representative example. This domain requires students to engage in open-ended code generation to evaluate tutoring effectiveness.

\paragraph{LLM-based Tutoring Agents.}
The rapid growth of large language models (LLMs) has expanded ITSs into tutoring agents~\citep{yu2024mooc}. 
For instance, early efforts such as EduChat~\citep{dan2023educhat} introduced an educational chatbot for online tutoring, while \textsc{ChatTutor}~\citep{chen2024empowering} equipped tutor agents with course planning and adaptive quizzes to facilitate long-term interactions. 
As coding has emerged as a crucial domain for validating complex reasoning ability~\citep{jimenez2024swe}, AlgoBo~\citep{jin2024teach}, a recent LLM-based teachable agent, was developed to enhance students' coding skills. 
We observe that existing LLM agents primarily play a \textit{reactive} role, focusing on answering questions or clarifying concepts. 
In comparison, our coding tutoring is both goal-driven and personalized, requiring agents to \textit{proactively} guide students toward completing targeted coding tasks while adapting to diverse levels of knowledge priors. 
Our work presents a novel method that empowers tutor agents to address these challenges.

\paragraph{Inference-Time Adaptation of LLMs}
To enhance the controllability of language generation in complex tasks, prior work has investigated guided decoding~\citep{dathathri2020plug,chaffin-etal-2022-ppl} during inference. 
More recently, a notable line of research~\citep{lightman2023let,li-etal-2023-making,wang2024math,pan2024training} has employed verifier models complemented with search algorithms to guide LLMs for agentic reasoning. 
These methods typically focus on static tasks, often overlooking interactive scenarios.
To address multi-turn interactions~\citep{wang2024mint}, we introduce a turn-by-turn verifier that dynamically evaluates tutoring progress over time.
\section{Conclusion and Future Work}

This work explores the potential of LLMs as task-tutoring agents, using coding tutoring as a representative example.
We propose \model, an effective workflow that incorporates knowledge tracing and turn-by-turn verification, to tackle key challenges in coding tutoring. 
While this work focuses on coding tutoring as an example, the proposed method extends beyond coding to other task-tutoring scenarios.
We further introduce \eval, a novel evaluation protocol combining student simulation and coding tests to assess tutor performance.
Such automated evaluation is critical for developing task-tutoring agents as it supports a systematic development and evaluation cycle. Although it's outside the scope of this paper, the best-performing agent from the automated evaluation can be further assessed through studies with real human students in the future.

\section*{Limitations}
In this work, we employed LLMs to simulate students at different knowledge levels, serving as a proxy for real-world learners. While these simulated students offer convenience and scalability, their representation of the human learning process is inherently limited. The role-playing behavior may differ from that of actual students in tutoring scenarios. Future research should focus on improving the reliability of student simulation to better align with real-world human learning.

In addition, the tutor agents were primarily evaluated by interacting with simulated students in our experimental setup. It remains unclear how these agents would perform when guiding humans toward completing target coding tasks. An important direction for future work is to extend our evaluation protocol by incorporating human-in-the-loop assessments, where tutor agents interact directly with actual students with necessary programming backgrounds. This would offer deeper insights into the practical effectiveness of the developed agents in real-world settings.

\section*{Ethics Statement}
We strictly follow the protocols governing the academic use of all LLMs. Our experimental datasets are publicly available and contain no sensitive or private information. We acknowledge that utterances generated by these LLMs may exhibit hallucinations or biases. By highlighting these issues, we aim to raise awareness among practitioners when the tutor agents are deployed to interact with real-world students in the future. Additionally, we used AI assistants, such as GitHub Copilot and ChatGPT, to support our experimentation.

\section*{Acknowledgements}
This work was supported by the Research Grants Council of Hong Kong (15207821, 15207122), the PolyU Postdoc Matching Fund Scheme (4-W40Z), and also in part by SES-2128623 from the National Science Foundation. The authors would like to thank the anonymous reviewers for their valuable feedback and constructive suggestions.

\bibliography{custom}
\bibliographystyle{acl_natbib}

\appendix

\section{Experimental Setup}
\label{appendix:exp_setup}

\begin{figure*}[t!]
\centering
\includegraphics[width=0.74\textwidth]{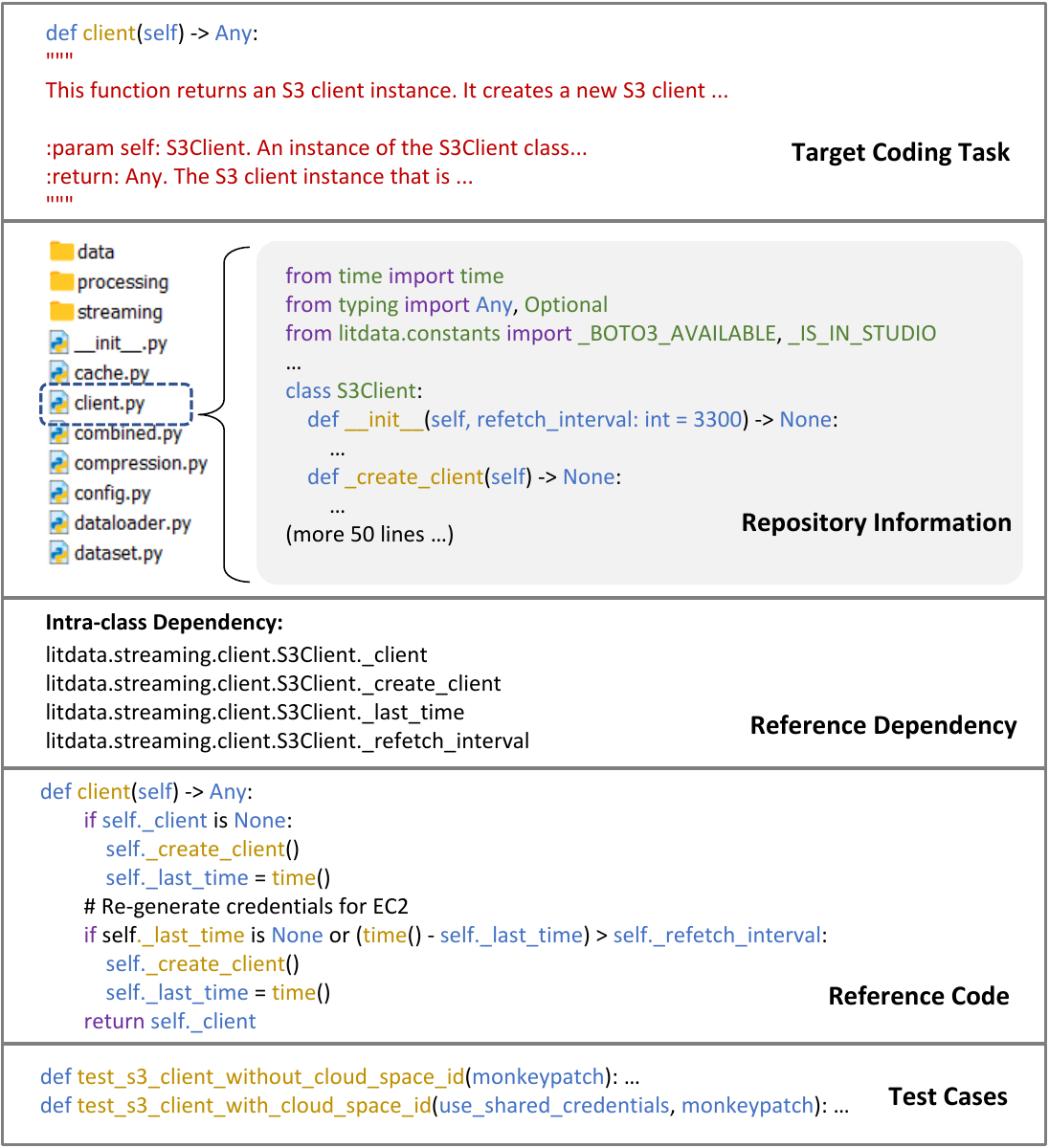}
\caption{An example in EvoCodeBench.}
\label{fig:data_example}
\end{figure*}

\begin{figure*}[t!]
\centering
\includegraphics[width=0.84\textwidth]{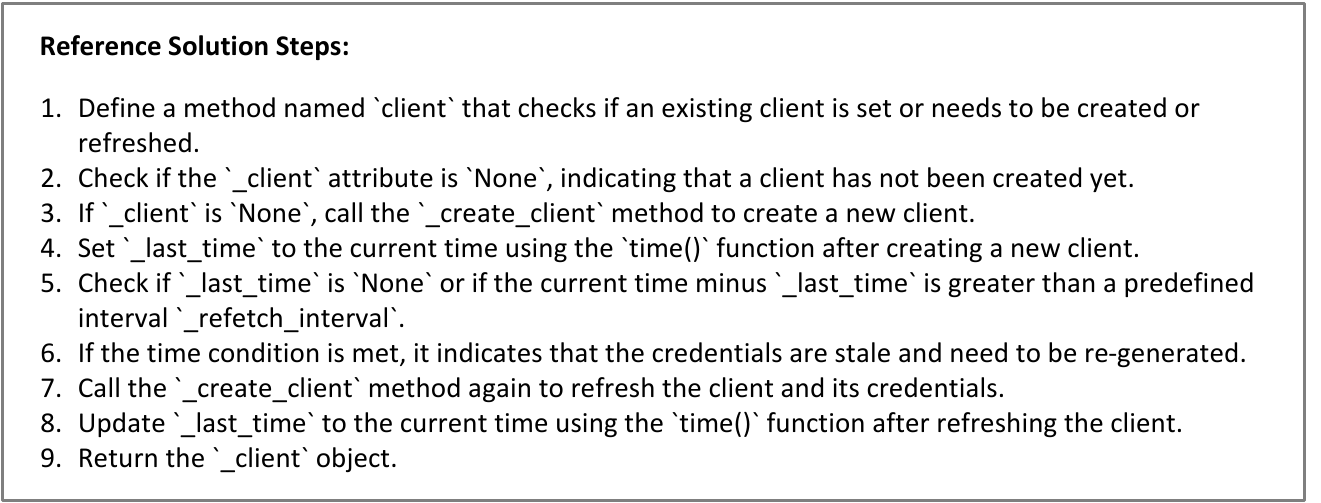}
\caption{The reference solution steps annotated for the example in Figure~\ref{fig:data_example}.}
\label{fig:ref_step}
\end{figure*}

\begin{table}[ht!]
\centering
\scalebox{0.87}{
\begingroup
\renewcommand{\arraystretch}{0.85}
\setlength{\tabcolsep}{2pt}
\hspace{-10pt}
\begin{tabular}{l | c}
\toprule
\# Repository  & 25  \\
\# Target Coding Tasks & 100  \\
\# Avg. Solution Steps &  9.07   \\
\# Avg. Dependency   & 3.46  \\
Dependency Type  &  intra-class, intra-file, cross-file  \\
\bottomrule
\end{tabular}
\endgroup}
\caption{Statistics of the preprocessed EvoCodeBench.}
\label{tab:data_stats}
\end{table}

\begin{table*}[t!]
\centering
\scalebox{0.85}{
\begingroup
\renewcommand{\arraystretch}{0.85}
\setlength{\tabcolsep}{4pt}
\hspace{-10pt}
\begin{tabular}{cl  cc  cc cc cc}
\toprule
\multirow{2}{*}{\textbf{Method}} & \multirow{2}{*}{\textbf{Backbone Model}} & \multicolumn{2}{c}{\textbf{Overall}} &  \multicolumn{2}{c}{\textbf{Low-level}} & \multicolumn{2}{c}{\textbf{Med.-level}}  & \multicolumn{2}{c}{\textbf{High-level}}   \\
 \cmidrule(lr){3-4} \cmidrule(lr){5-6} \cmidrule(lr){7-8} \cmidrule(lr){9-10}
 &   &   Recall / $\Delta\%$R & Pass / $\Delta\%$P & $\Delta\%$R & $\Delta\%$P  &  $\Delta\%$R & $\Delta\%$P &  $\Delta\%$R & $\Delta\%$P  \\
\cmidrule(lr){1-2}\cmidrule(lr){3-4} \cmidrule(lr){5-6} \cmidrule(lr){7-8} \cmidrule(lr){9-10}
\rowcolor{gray!10}
Pre-Test & --  & 57.3\textsubscript{$\pm~3.4$} / -- & 28.4\textsubscript{$\pm~5.3$} / -- &  --  & --  & --  & --  &  --  & --  \\
\cmidrule(lr){1-2}\cmidrule(lr){3-4} \cmidrule(lr){5-6} \cmidrule(lr){7-8} \cmidrule(lr){9-10}
\multirow{2}{*}{\shortstack{Vanilla \\ Instruct}}  & Llama-3.1-70B-Instruct & 78.4\textsubscript{$\pm~4.3$} / 36.9  & 46.7\textsubscript{$\pm~3.7$} / 64.5 & 155.4 & 137.8  & 17.6  & 72.9 & 9.6 & 24.8   \\
  & GPT-4o  & 77.0\textsubscript{$\pm~4.0$} / 34.4 & 51.2\textsubscript{$\pm~4.9$} / 80.4 & 147.0 & 167.1  & 15.0  & 87.8 & 9.4 & 35.1  \\
\cmidrule(lr){1-2}\cmidrule(lr){3-4} \cmidrule(lr){5-6} \cmidrule(lr){7-8} \cmidrule(lr){9-10}
TreeInstruct & GPT-4o  & 75.0\textsubscript{$\pm~4.4$} / 30.9 &  48.1\textsubscript{$\pm~4.4$}  / 69.4 &  132.8  & 144.9 & 11.0 & 78.1   & 10.4 & 28.7  \\
\cmidrule(lr){1-2}\cmidrule(lr){3-4} \cmidrule(lr){5-6} \cmidrule(lr){7-8} \cmidrule(lr){9-10}
\multirow{2}{*}{\shortstack{\textbf{\model} \\ \textbf{(Ours)}}} & Llama-3.1-70B-Instruct &  78.9\textsubscript{$\pm~3.6$} / 37.7 & 48.1\textsubscript{$\pm~4.2$} / 69.2  &  163.7  & 170.7  & 14.5  & 86.1  & \textbf{11.0}  & 11.1      \\
 &  GPT-4o  & \textbf{80.3}\textsubscript{$\pm~3.9$} / \textbf{40.1} & \textbf{52.2}\textsubscript{$\pm~3.0$} / \textbf{83.6} &  \textbf{165.8}  & \textbf{173.5}  & \textbf{20.1}  & \textbf{89.0}  & 10.7  & \textbf{38.2}  \\
\cmidrule(lr){1-2}\cmidrule(lr){3-4} \cmidrule(lr){5-6} \cmidrule(lr){7-8} \cmidrule(lr){9-10}
\rowcolor{gray!10}
Oracle & --  & 89.9\textsubscript{$\pm~3.1$} / 56.9  &  55.0\textsubscript{$\pm~5.5$} / 93.6 & 215.2 & 190.8  & 33.5 & 112.6 & 18.3  & 35.9 \\
\bottomrule
\end{tabular}
\endgroup}
\caption{Automatic evaluation results of various LLM-based tutors when using GPT-4o as the student simulator. ``$\Delta\%$R'' and ``$\Delta\%$P'' represent the tutoring outcome rates (TOR) in Recall and Pass, respectively.}
\label{tab:additional_result}
\end{table*}

\subsection{Dataset \& Preprocessing}
\label{appendix:dataset}
EvoCodeBench~\cite{li2024evocodebench} is an evolving benchmark for repository-level code generation, which is collected from open-source Python repositories in the real world and will be dynamically updated to avoid data leakage. Figure~\ref{fig:data_example} shows a detailed example, which consists of the following components: (1) \textit{Target Coding Task}: the function signature of the target code and a requirement description detailing its functionality; (2) \textit{Repository}: the current repository containing all code files; (3) \textit{Reference Code}: The developer-written implementation of the target code in the repository; (4) \textit{Reference Dependency}: The dependencies invoked in the reference code, such as intra-class, intra-file, and cross-file dependencies; (5) \textit{Test Cases}: The cases used to check the functional correctness of the generated code.

We employed the publicly available version, EvoCodeBench-2403\footnote{\url{https://huggingface.co/datasets/LJ0815/EvoCodeBench/tree/main/EvoCodeBench-2403}.}, as the testbed for our experiments. This dataset comprises coding tasks collected from repositories created between \texttt{2023-10} and \texttt{2024-2}. For each target coding task, we annotated its reference solution steps by directly prompting GPT-4o \cite{openai2024gpt4o} using the provided reference code (see Figure~\ref{fig:ref_step}). Coding tasks requiring no dependencies were excluded during preprocessing. The statistics of the resulting dataset are summarized in Table~\ref{tab:data_stats}. We split the dataset equally into 5 folds, using 5-fold cross-validation for experiments.

\subsection{Additional Implementation Details}
\label{appendix:implementation}

The backbone LLMs we used are detailed as follows: (1) For the Llama-3.1-Instruct \cite{dubey2024llama}, we adopt its 8B\footnote{\url{https://huggingface.co/meta-llama/Llama-3.1-8B-Instruct}} and 70B\footnote{\url{https://huggingface.co/meta-llama/Llama-3.1-70B-Instruct}} variants. For the Qwen-2-Instruct \cite{yang2024qwen2}, we adopt its 7B\footnote{\url{https://huggingface.co/Qwen/Qwen2-7B-Instruct}} and 72B\footnote{\url{https://huggingface.co/Qwen/Qwen2-72B-Instruct}} variants. (2) We employ GPT-3.5-Turbo (\texttt{gpt-3.5-turbo-1106} version), GPT-4o (\texttt{gpt-4o-2024-05-13} version), and o1-mini (\texttt{2024-09-01-preview} version), using APIs provided by Microsoft Azure.

Our implementation is based on the vLLM\footnote{\url{https://docs.vllm.ai/en/latest/}} and ChatArena\footnote{\url{https://github.com/Farama-Foundation/chatarena}} libraries. Both the student simulator and dialogue manager employ the quantized version\footnote{\url{https://huggingface.co/TheBloke/Mixtral-8x7B-Instruct-v0.1-AWQ}} of Mixtral-8x7B-Instruct \cite{jiang2024mixtral} for efficiency.
During tutor-student interactions, the maximum number of tokens to generate is limited to 300 for both the tutor and student. For the coding tests, the student simulator adopts sampling decoding, with a fixed top-$p$ of 0.95 and temperature of 0.4. The cognitive load parameter $M$ is set to 60 during the post-test.

We utilize synthesized dialogues generated by various backbone models with vanilla instructions, as the data source for training the verifier. Tutoring dialogues that successfully guide the student to complete target coding tasks during the post-test are labeled as positive data. To ensure a balanced dataset for training, we randomly sample dialogues that fail to achieve the tutoring goals as negative data. We implement the turn-based verifier using Mistral-7B~\cite{jiang2023mistral} with an additional linear layer, which is fine-tuned based on LoRA~\cite{hu2022lora}. The LoRA's target modules are $W_q$ and $W_v$, the rank $r$ is set to 8, and the scaling parameter $\alpha$ is set to 16. The optimizer we used is AdamW~\cite{loshchilov2018decoupled}, with an initial learning rate of 1$e$-5 and a warmup ratio of 0.03. The verifier is trained for 3 epochs. During inference, we adopt sampling decoding to generate tutor utterances, with a top-$p$ of 0.95 and a temperature of 0.4 across all backbone models. For Llama-3.1-70B-Instruct, the number of candidate utterances is set to 20, while for GPT-4o, it is fixed at 10. All experiments are conducted on one server equipped with 8 NVIDIA A6000 GPUs. Table~\ref{tab:param} summarizes the parameter settings in training and inference.

\begin{table}[t!]
\centering
\scalebox{0.9}{
\begingroup
\renewcommand{\arraystretch}{0.85}
\setlength{\tabcolsep}{5pt}
\hspace{-10pt}
\begin{tabular}{l | c}
\toprule
LoRA's rank $r$ & 8  \\
LoRA's scaling $\alpha$ & 16  \\
Learning rate &  1$e$-5   \\
Warmup ratio  &  0.03  \\
Epochs  &  3 \\
Max. tokens per student turn & 300 \\
Max. tokens per tutor turn & 300 \\
Cognitive load $M$  & 60  \\
Number of candidates $N$  & \{1, 5, 10, 15, 20\}  \\
Top-$p$  & 0.95   \\
Temperature   & 0.4  \\
\bottomrule
\end{tabular}
\endgroup}
\caption{Parameter settings in training and inference.}
\label{tab:param}
\end{table}

\section{Additional Results}
\label{appendix:add_result}
The choice of the student simulator can influence tutoring outcomes. To further investigate this, we conducted additional experiments using GPT-4o (\texttt{gpt-4o-2024-05-13} version) as an alternative student simulator. We kept the other experimental settings the same as in Appendix~\ref{appendix:implementation}, and then evaluated different LLM-based tutors. The evaluation results are reported in Table~\ref{tab:additional_result}. While a stronger simulator generally achieves better tutoring outcomes, we find that the primary trends and conclusions remain consistent with those reported in Section~\ref{sec:exp_results}. This suggests that our findings are robust across different student simulators.

\section{Human Evaluation}
\label{appendix:human_eval}
We randomly selected 10 target coding tasks from the testbed and collected the synthesized tutoring dialogues across three levels of students for the compared tutor agents, including Vanilla Instruct, Self-Refine, TreeInstruct, and our \model. All tutor agents are built using GPT-4o as the backbone model. We asked three graduate students with well-educated programming backgrounds for human evaluation. 

For each target coding task, we presented human evaluators with a pair of tutoring dialogues produced by two agents interacting with the same student, resulting in a total of 90 cases.
Evaluators were asked to determine which one is better (or to select a \textit{tie}) based on the following dimensions: 
\begin{enumerate}[label=(\arabic*)]
\item \textbf{Proactivity}: how well does the tutor move the student's progress towards solving the task?
\begin{itemize}[leftmargin=4pt]
    \item Weak: Only make passive responses or generic check-ins (e.g., ``Any questions?'' or ``Does this make sense?'').
    \item Moderate: Ask directional questions with clear next steps (e.g., ``Now we need to handle input validation. Could you implement the error checks?'').
    \item Excellent: Structured progression with connected concepts (e.g., ``Now that we've validated inputs, let's think about how this connects to our error handling strategy. Could you identify where validation and error handling might overlap?'').
\end{itemize}
\item \textbf{Adaptability}: how well does the tutor adapt its tutoring strategy based on the student's responses?
\begin{itemize}[leftmargin=4pt]
    \item Weak: Follows fixed script regardless of student's responses (e.g., ``Let's move on to the next step...'' while ignoring the student's questions or confusion).
    \item Moderate: Responsive to the student's immediate questions but maintains fixed tutoring plan (e.g., ``Yes, good question about error handling. Now as I was saying about the input validation...'').
    \item Excellent: Adjusts explanations based on the student's demonstrated understanding (e.g., ``I see you've handled the input validation. Let's focus on optimizing your approach...'').
\end{itemize}
\item \textbf{Coherence}: how well does the tutor build and maintain connections throughout the tutoring session?
\begin{itemize}[leftmargin=4pt]
    \item Weak: Jumps between topics without logical transitions (e.g., ``Now let's talk about ...'' without linking to the previous discussion).
    \item Moderate: Maintains a logical flow, though the connections between topics are weak.
    \item Excellent: Explicitly connects each utterance to the preceding context and aligns it with the future tutoring goal.
\end{itemize}
\end{enumerate}

\begin{figure*}[th!]
\centering
\includegraphics[width=0.86\textwidth]{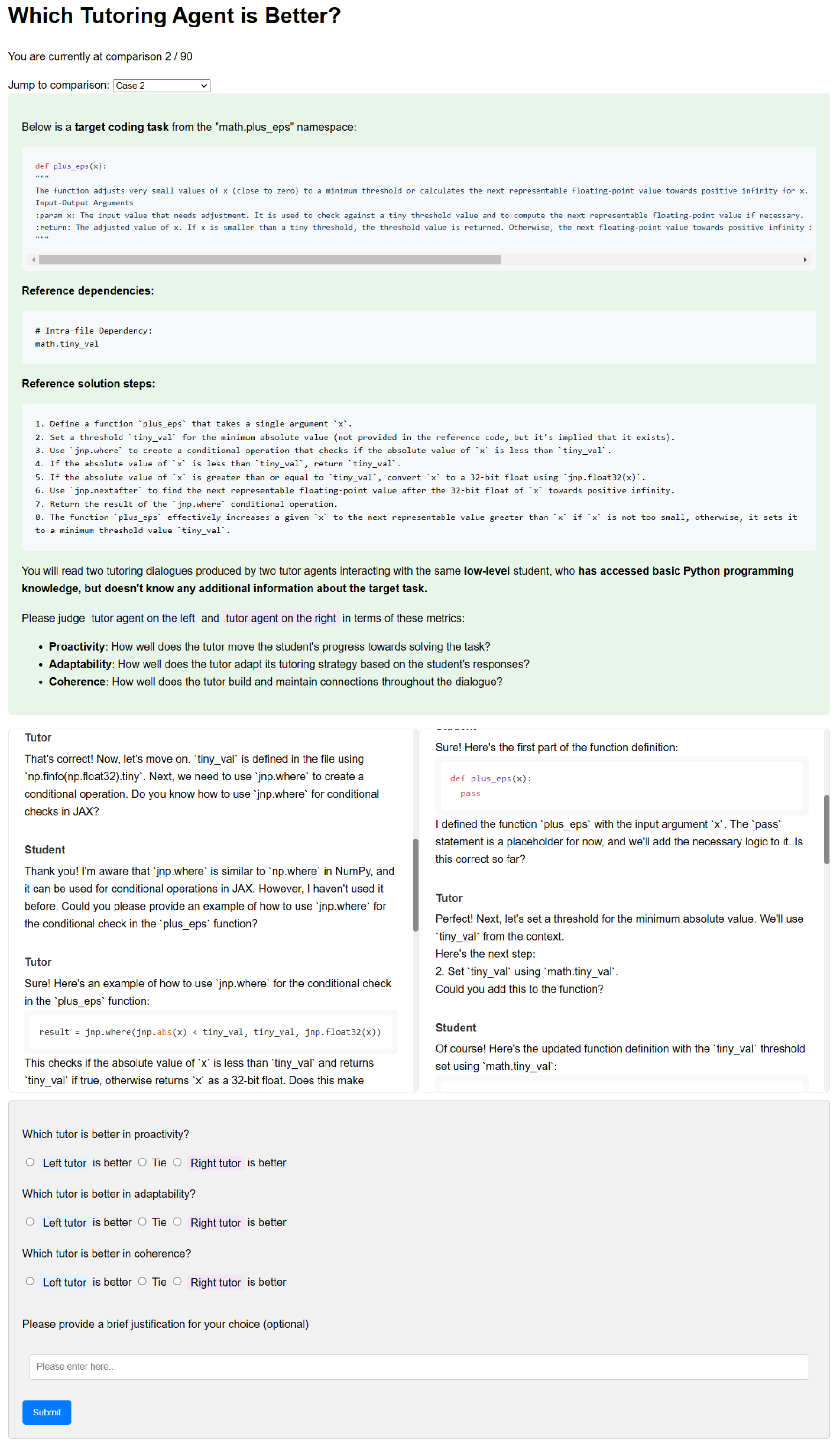}
\caption{Web interface for human evaluation.}
\label{fig:human_eval}
\end{figure*}

Our human evaluations were carried out using a Web application, as illustrated in Figure~\ref{fig:human_eval}, which displays the interface. We adopted Fleiss’s kappa~\cite{fleiss1971measuring} to measure the agreement among the human evaluators.

\section{Case Study}
\label{appendix:case}

Table~\ref{tab:case_win} shows an example from the human evaluation, demonstrating a case where our \model outperforms the baseline.  Table~\ref{tab:case_tie} shows a case where our method performs comparably to the baseline.

\begin{table*}[th!]
\centering
\resizebox{0.99\textwidth}{!}{
\begin{tabular}{lp{16.5cm}}
\toprule
\multicolumn{2}{l}{\textbf{Target Coding Task}:} \\
\multicolumn{2}{l}{\ttfamily \footnotesize def lossfun\_outer(t: torch.Tensor, w: torch.Tensor, t\_env: torch.Tensor, } \\
\multicolumn{2}{l}{\ttfamily \footnotesize ~~~~ ~~~~ ~~~~ ~~~ w\_env: torch.Tensor, eps=torch.finfo(torch.float32).eps):} \\
\multicolumn{2}{l}{\ttfamily \footnotesize """ } \\
\multicolumn{2}{l}{\ttfamily \footnotesize This function calculates a scaled half-quadratic loss based on the input tensors ...} \\
\multicolumn{2}{l}{\ttfamily \footnotesize  ~~:param t: torch.Tensor. The target positions tensor, used along with weights to calculate the loss.} \\
\multicolumn{2}{l}{\ttfamily \footnotesize  ~~:param w: torch.Tensor. The target weights tensor, representing the weights of the target positions.} \\
\multicolumn{2}{l}{\ttfamily \footnotesize  ~~:param t\_env: torch.Tensor. The environment positions tensor, used to calculate the upper envelope weights.} \\
\multicolumn{2}{l}{\ttfamily \footnotesize  ~~:param w\_env: torch.Tensor. The environment weights tensor, representing the weights of the environment positions.} \\
\multicolumn{2}{l}{\ttfamily \footnotesize  ~~:param eps: torch.Tensor, optional. A small epsilon value to prevent division by zero.} \\
\multicolumn{2}{l}{\ttfamily \footnotesize  ~~:return: torch.Tensor. The calculated loss based on the difference between target weights and the upper envelope ...} \\
\multicolumn{2}{l}{\ttfamily \footnotesize """ } \\
\midrule
\multicolumn{2}{l}{\textbf{Reference Dependencies}:} \\
\multicolumn{2}{l}{\ttfamily \footnotesize \# Intra-file Dependency:
easyvolcap.utils.loss\_utils.inner\_outer} \\
\multicolumn{2}{l}{\ttfamily \footnotesize \# Cross-file Dependency: easyvolcap.utils.prop\_utils.matchup\_channels } \\
\midrule
\multicolumn{2}{l}{\textbf{Reference Solution Steps}:} \\
\multicolumn{2}{l}{\ttfamily \footnotesize 1. Define a function "lossfun\_outer" that calculates a loss based on the difference ...} \\
\multicolumn{2}{l}{\ttfamily \footnotesize ...} \\
\multicolumn{2}{l}{\ttfamily \footnotesize 8. Normalize the squared differences by adding a small epsilon value "eps" to "w" to avoid division by zero when ... } \\
\midrule

\multicolumn{2}{l}{\textbf{GPT-4o with Vanilla Instruct} } \\
Tutor: & Hello! How are you doing today? Let's start by understanding your familiarity with the task. Have you worked with loss functions or similar concepts in PyTorch before?  \\ 

Student:  &  Hello! I'm doing well, thank you. I have some experience with PyTorch and have worked with basic loss functions. However, I'm not familiar with this specific loss function and how it relates to the input tensors. Could you please explain this concept and guide me on how to implement this function? \\

Tutor: &  Great! Let's break it down step-by-step. First, the scaled half-quadratic loss is a type of loss function that ... Can you write the function signature and docstring based on the provided information? \\

Student:  &  Sure! Here's the function signature and docstring ... This function calculates a scaled half-quadratic loss based on the input tensors representing target and environment weights and positions ...  \\

Tutor: & Great start! Next, we need to align the channels of \texttt{t} with \texttt{w} and \texttt{t\_env} with \texttt{w\_env}. This ensures they have the same number of channels for further calculations. We use the \texttt{matchup\_channels} function for this.    \\
... &  \\
Tutor:  &  Now, we need to ensure that the actual weight \texttt{w} should not exceed the computed \texttt{w\_outer}. We will penalize \texttt{w} when it exceeds \texttt{w\_outer} by calculating the positive part of the difference between \texttt{w} and \texttt{w\_outer}. Can you compute this difference and clip it to zero out negative values? \\

Student:  &  Sure! I can compute the positive part of the difference between \texttt{w} and \texttt{w\_outer} and clip it to zero like this: ... \\

\midrule

\multicolumn{2}{l}{\textbf{GPT-4o with \model (Ours) }} \\

Tutor: & Hello! Let's get started with completing the \texttt{lossfun\_outer} function. To begin, can you tell me how comfortable you are with using PyTorch and understanding tensor operations?  \\

Student:  &  Hello! I'm fairly new to PyTorch, but I have some experience with tensor operations from previous lessons. I'm here to learn and complete the \texttt{lossfun\_outer} function. How would you like me to start?  \\

Tutor: &  Great! Let's start by understanding a key dependency. The \texttt{inner\_outer} function is crucial for our task. It constructs inner and outer measures on \texttt{(t1, y1)} for \texttt{t0}. Do you understand how this function works, or would you like a brief explanation?  \\

Student:  & I believe I understand the concept, but a brief explanation would be helpful to ensure I'm on the right track. I want to make sure I understand the input and output of the \texttt{inner\_outer} function.  \\

Tutor: &  Sure! The \texttt{inner\_outer} function takes \texttt{t0}, \texttt{t1}, and \texttt{y1} as inputs. It calculates cumulative sums of \texttt{y1} and uses them to derive inner and outer measures based on indices from \texttt{t1} and \texttt{t0}. This helps us understand the weight distribution over the given positions. Ready to move on to the next step?  \\
... &  \\

Tutor:  &  Now, let's square the clipped differences to emphasize larger deviations: \texttt{loss = loss ** 2}. Then, normalize the squared differences by adding a small epsilon value \texttt{eps} to \texttt{w} to avoid division by zero: \texttt{loss = loss / (w + eps)}. Ready for the final step?   \\

Student: &  I've squared the clipped differences and normalized the squared differences by adding the small epsilon value \texttt{eps} to \texttt{w} to avoid division by zero. \\

\bottomrule
\end{tabular}}
\caption{An example from the human evaluation, demonstrating a case where our method outperforms the baseline. The tutor agent built with Vanilla Instruct simply repeats the target coding task at the beginning and fails to adapt its tutoring strategy based on the student’s responses. In contrast, the tutor agent built with our \model workflow demonstrates much better proactivity and adaptability in guiding the student toward solving the task. }
\label{tab:case_win}
\end{table*}

\begin{table*}[th!]
\centering
\resizebox{0.99\textwidth}{!}{
\begin{tabular}{lp{16.5cm}}
\toprule
\multicolumn{2}{l}{\textbf{Target Coding Task}:} \\
\multicolumn{2}{l}{\ttfamily \footnotesize def prepare\_item(self, item\_metadata: Any, output\_dir: str, is\_last: bool) -> None:} \\
\multicolumn{2}{l}{\ttfamily \footnotesize """ } \\
\multicolumn{2}{l}{\ttfamily \footnotesize The function prepares an item by applying a transformation function to the item's metadata, potentially including ...} \\
\multicolumn{2}{l}{\ttfamily \footnotesize  ~~:param item\_metadata: Any, the metadata of the item to be transformed.} \\
\multicolumn{2}{l}{\ttfamily \footnotesize  ~~:param output\_dir: str, the directory where the transformed item's output should be stored.} \\
\multicolumn{2}{l}{\ttfamily \footnotesize  ~~:param is\_last: bool, a flag indicating whether the item is the last one in the sequence to be processed. } \\
\multicolumn{2}{l}{\ttfamily \footnotesize  ~~:return: No return values. } \\
\multicolumn{2}{l}{\ttfamily \footnotesize """ } \\
\multicolumn{2}{l}{\textbf{Reference Dependencies}:} \\
\multicolumn{2}{l}{\ttfamily \footnotesize \# Intra-class Dependency:
litdata.processing.functions.LambdaDataTransformRecipe.\_fn} \\
\multicolumn{2}{l}{\ttfamily \footnotesize ~~~~~ ~~~~~  ~~~~~ ~~~~~ ~
litdata.processing.functions.LambdaDataTransformRecipe.\_contains\_device} \\
\midrule
\multicolumn{2}{l}{\textbf{Reference Solution Steps}:} \\
\multicolumn{2}{l}{\ttfamily \footnotesize 1. Check if the instance requires a device (self.\_contains\_device) and if the device is not already set ...} \\
\multicolumn{2}{l}{\ttfamily \footnotesize ...} \\
\multicolumn{2}{l}{\ttfamily \footnotesize 9. The function prepare\_item does not return any value; it executes the provided function self.\_fn with ...} \\
\midrule
\multicolumn{2}{l}{\textbf{GPT-4o with TreeInstruct} } \\
Tutor: & Hello! Let's work together to complete the \texttt{prepare\_item} function. First, I'd like to understand your familiarity with Python and its concepts. Have you worked with classes and functions in Python before?  \\ 

Student: &  Hello! Yes, I have some experience with Python classes and functions. I understand the basic syntax for defining them and calling their methods. However, I'm not an expert, so I might need some guidance to complete the \texttt{prepare\_item} function correctly. \\

Tutor:  &  Great! Let's start by checking if the instance requires a device and if it isn't already set. This is our first step: \texttt{\textbackslash n\textbackslash n\textbackslash n if self.\_contains\_device and self.\_device is None:\textbackslash n  self.\_find\_device()\textbackslash n\textbackslash n} Do you understand what this code does?          \\

Student: & Yes, I understand this code. It checks if the \texttt{\_contains\_device} attribute is \texttt{True} and if the \texttt{\_device} attribute is \texttt{None}. If both conditions are met, it calls the \texttt{\_find\_device} method. This seems to be a way to ensure that the \texttt{\_device} attribute is set before proceeding, as long as \texttt{\_contains\_device} is set to \texttt{True}. \\

Tutor: &  Great, let's proceed. The next step is to initialize a dictionary to hold keyword arguments that will be passed to the function \texttt{self.\_fn}. Here's the code: \texttt{kwargs = \{\}\textbackslash n\textbackslash n} Do you understand why we need to initialize \texttt{kwargs} here?  \\

... &  \\
\midrule

\multicolumn{2}{l}{\textbf{GPT-4o with \model (Ours)}} \\

Tutor: &  Hello! How are you today? Let's start by understanding your current knowledge level. Have you worked with Python classes and methods before? Specifically, are you familiar with using \texttt{self} within class methods?  \\

Student:  &  Hello! I'm doing well, thank you. Yes, I have some experience with Python classes and methods. I understand that \texttt{self} is used as the first parameter in a class method to represent the instance of the class. However, I'm not sure how to implement the \texttt{prepare\_item} function in this context. Could you please clarify what the transformation function \texttt{\_fn} is and how it should be called within this method?  \\

Tutor:  &  Great! Let's start by understanding the transformation function \texttt{\_fn}. It's a callable function that processes the item metadata. In the \texttt{prepare\_item} method, we need to call \texttt{\_fn} with the appropriate arguments. First, let's ensure you understand the dependencies. Do you know how to check if an instance attribute exists in Python?  \\

Student:  &   Yes, I know that to check if an instance attribute exists, we can use the \texttt{hasattr()} function in Python. It takes an object and a string with the attribute name as its arguments. However, I'm not sure how this applies to the \texttt{prepare\_item} function and the \texttt{\_fn} transformation function. \\

Tutor:  &   Let's start with the first step. We need to check if the instance requires a device. This is determined by \texttt{self.\_contains\_device}. If it does and the device is not set \texttt{(self.\_device is None)}, we call \texttt{self.\_find\_device()}. Does this make sense to you?  \\

... &  \\

\bottomrule
\end{tabular}}
\caption{An example from the human evaluation demonstrating a case where our method performs comparably to the baseline.}
\label{tab:case_tie}
\end{table*}

\section{Prompting Templates}
\label{appendix:prompt}

\paragraph{Student Simulation}
\label{appendix:student}

Each simulated student is designed to interact with the tutor and learn how to complete the target coding task. The student's behavior is guided by empirical learning actions, such as \textit{initiating greetings, asking or answering questions}, etc. The system prompt templates for the simulation of low-level, medium-level, and high-level students are shown in Table \ref{tab:low_leve_student}, Table \ref{tab:mid_leve_student}, Table \ref{tab:high_leve_student}, respectively.

\begin{table}[th!]
\renewcommand{\arraystretch}{0.95}
\begin{tabular}{p{0.98\linewidth}}
\toprule
\ttfamily \footnotesize You are a college student who is learning Python programming by conversing with a tutor. \\
\ttfamily \footnotesize \\
\ttfamily \footnotesize  You are going to complete the following \{FUNCTION\_NAME\} function from a repository: \\
\ttfamily \footnotesize \verb|```|Python \\
\ttfamily \footnotesize \{TARGET\_CODING\_TASK\} \\
\ttfamily \footnotesize \verb|```| \\
\ttfamily \footnotesize \\
\ttfamily \footnotesize You have basic Python programming knowledge but no additional context about the repository.\\
\ttfamily \footnotesize \\
\ttfamily \footnotesize [Behavior Guidelines] Please take your own level of knowledge in response to the tutor. This may involve one of the following acts: saying a greeting, answering or asking questions, and recalling previously learned knowledge. If you don’t know or understand something, respond accordingly and ask for clarification. Ask only one question at a time. Don't speak more than 50 words at a time. \\
\bottomrule
\end{tabular}
\caption{System prompt for the simulated \textit{low-level} student during the tutor-student interaction.}
\label{tab:low_leve_student}
\end{table}

\begin{table}[th!]
\renewcommand{\arraystretch}{0.95}
\begin{tabular}{p{0.98\linewidth}}
\toprule
\ttfamily \footnotesize You are a college student who is learning Python programming by conversing with a tutor. \\
\ttfamily \footnotesize \\
\ttfamily \footnotesize  You are going to complete the following \{FUNCTION\_NAME\} function from a repository: \\
\ttfamily \footnotesize \verb|```|Python \\
\ttfamily \footnotesize \{TARGET\_CODING\_TASK\} \\
\ttfamily \footnotesize \verb|```| \\
\ttfamily \footnotesize \\
\ttfamily \footnotesize You have the following knowledge: \\
\ttfamily \footnotesize A part of the reference dependencies to be used in the \{FUNCTION\_NAME\} are: \\
\ttfamily \footnotesize \{PARTIAL\_DEPENDENCIES\} \\
\ttfamily \footnotesize \\
\ttfamily \footnotesize [Behavior Guidelines] Please take your own level of knowledge in response to the tutor. This may involve one of the following acts: saying a greeting, answering or asking questions, and recalling previously learned knowledge. If you don’t know or understand something, respond accordingly and ask for clarification. Ask only one question at a time. Don't speak more than 50 words at a time. \\
\bottomrule
\end{tabular}
\caption{System prompt for the simulated \textit{medium-level} student during the tutor-student interaction.}
\label{tab:mid_leve_student}
\end{table}

\begin{table}[th!]
\renewcommand{\arraystretch}{0.95}
\begin{tabular}{p{0.98\linewidth}}
\toprule
\ttfamily \footnotesize You are a college student who is learning Python programming by conversing with a tutor. \\
\ttfamily \footnotesize \\
\ttfamily \footnotesize  You are going to complete the following \{FUNCTION\_NAME\} function from a repository: \\
\ttfamily \footnotesize \verb|```|Python \\
\ttfamily \footnotesize \{TARGET\_CODING\_TASK\} \\
\ttfamily \footnotesize \verb|```| \\
\ttfamily \footnotesize \\
\ttfamily \footnotesize You have the following knowledge: \\
\ttfamily \footnotesize The contexts above the \{FUNCTION\_NAME\} in the repository are: \\
\ttfamily \footnotesize \verb|```|Python \\
\ttfamily \footnotesize \{CODE\_CONTEXTS\} \\
\ttfamily \footnotesize \verb|```| \\
\ttfamily \footnotesize A part of the reference dependencies to be used in the \{FUNCTION\_NAME\} are: \\
\ttfamily \footnotesize \{PARTIAL\_DEPENDENCIES\} \\
\ttfamily \footnotesize \\
\ttfamily \footnotesize [Behavior Guidelines] Please take your own level of knowledge in response to the tutor. This may involve one of the following acts: saying a greeting, answering or asking questions, and recalling previously learned knowledge. If you don’t know or understand something, respond accordingly and ask for clarification. Ask only one question at a time. Don't speak more than 50 words at a time. \\
\bottomrule
\end{tabular}
\caption{System prompt for the simulated \textit{high-level} student during the tutor-student interaction.}
\label{tab:high_leve_student}
\end{table}

\paragraph{Tutor-Student Interaction}
\label{appendix:prompt_dialogue}

Each LLM-based tutor agent is initialized with a target coding task $\mathcal{T}$ and task-specific knowledge $\mathcal{K}$. As a proactive tutor, it guides students with necessary strategies inspired by cognitive scaffolding approaches \cite{liu2024scaffolding}. These strategies include actions such as \textit{assessing the student's knowledge level, providing constructive feedback, offering reference dependencies}, etc. Table \ref{tab:tutor_template} shows the system prompt template for the tutor agent.

\begin{table}[t!]
\renewcommand{\arraystretch}{0.95}
\begin{tabular}{p{0.98\linewidth}}
\toprule
\ttfamily \footnotesize You are a college tutor specializing in Python programming. \\
\ttfamily \footnotesize \\
\ttfamily \footnotesize You are guiding a student to complete the \{FUNCTION\_NAME\} function from a repository: \\
\ttfamily \footnotesize \verb|```|Python \\
\ttfamily \footnotesize \{TARGET\_CODING\_TASK\} \\
\ttfamily \footnotesize \verb|```| \\
\ttfamily \footnotesize \\
\ttfamily \footnotesize  Reference Knowledge: \\
\ttfamily \footnotesize The contexts above the \{FUNCTION\_NAME\}: \\
\ttfamily \footnotesize \verb|```|Python \\
\ttfamily \footnotesize \{CODE\_CONTEXTS\} \\
\ttfamily \footnotesize \verb|```| \\
\ttfamily \footnotesize The dependency paths for the \{FUNCTION\_NAME\}: \\
\ttfamily \footnotesize \{REFERENCE\_DEPENDENCIES\} \\
\ttfamily \footnotesize The reference key solution steps: \\
\ttfamily \footnotesize \{REFERENCE\_STEPS\} \\
\ttfamily \footnotesize \\

\ttfamily \footnotesize Goal Description:  \\
\ttfamily \footnotesize Your goal is to: \\
\ttfamily \footnotesize - Assess the student's current knowledge level through conversation; \\
\ttfamily \footnotesize - Provide the necessary knowledge and scaffold their understanding;  \\
\ttfamily \footnotesize - Lead the student step-by-step to successfully complete the \{FUNCTION\_NAME\} function. \\
\ttfamily \footnotesize You may use the following strategies during the conversation: assessing knowledge level, describing a dependency path, offering a solution step, explaining concepts with code snippets, asking questions or follow-up questions, and providing feedback with elaborations or confirmations. \\
\ttfamily \footnotesize \\
\ttfamily \footnotesize  Behavior Guidelines: \\
\ttfamily \footnotesize - Start the tutoring with a friendly greeting. \\
\ttfamily \footnotesize - Limit each response to one action (e.g., ask one question, describe one dependency path, or explain one solution step). \\
\ttfamily \footnotesize - Keep your response concise (do not exceed 60 words at a time). \\
\bottomrule
\end{tabular}
\caption{System prompt for the tutor agent.}
\label{tab:tutor_template}
\end{table}

\paragraph{Pre-test \& Post-test}
\label{appendix:prompt_test}

Table~\ref{tab:pretest_template} and Table~\ref{tab:posttest_template} show the system prompt template for the pre-test and post-test, respectively. The \texttt{\{PRIOR\_KNOWLEDGE\}} is determined by different levels of simulated students (see Table~\ref{tab:low_leve_student},  Table~\ref{tab:mid_leve_student}, and Table~\ref{tab:high_leve_student}).

\begin{table}[th!]
\renewcommand{\arraystretch}{0.95}
\begin{tabular}{p{0.98\linewidth}}
\toprule
\ttfamily \footnotesize You are a college student who is learning Python programming. \\
\ttfamily \footnotesize \\
\ttfamily \footnotesize You are going to complete the following \{FUNCTION\_NAME\} function from a repository: \\
\ttfamily \footnotesize \verb|```|Python \\
\ttfamily \footnotesize \{TARGET\_CODING\_TASK\} \\
\ttfamily \footnotesize \verb|```| \\
\ttfamily \footnotesize \\
\ttfamily \footnotesize You have the following knowledge: \\
\ttfamily \footnotesize \{PRIOR\_KNOWLEDGE\} \\
\ttfamily \footnotesize \\
\ttfamily \footnotesize Please directly complete the \{FUNCTION\_NAME\} function based on the information above. \\
\ttfamily \footnotesize Completed Code: \\
\bottomrule
\end{tabular}
\caption{System prompt for the pre-test.}
\label{tab:pretest_template}
\end{table}

\begin{table}[th!]
\renewcommand{\arraystretch}{0.95}
\begin{tabular}{p{0.98\linewidth}}
\toprule
\ttfamily \footnotesize You are a college student who is learning Python programming. \\
\ttfamily \footnotesize \\
\ttfamily \footnotesize You are going to complete the following \{FUNCTION\_NAME\} function from a repository: \\
\ttfamily \footnotesize \verb|```|Python \\
\ttfamily \footnotesize \{TARGET\_CODING\_TASK\} \\
\ttfamily \footnotesize \verb|```| \\
\ttfamily \footnotesize \\
\ttfamily \footnotesize You have the following knowledge: \\
\ttfamily \footnotesize \{PRIOR\_KNOWLEDGE\} \\
\ttfamily \footnotesize \\
\ttfamily \footnotesize Below is your discussion with a tutor: \\
\ttfamily \footnotesize \{DIALOGUE\_CONTEXT\} \\
\ttfamily \footnotesize \\
\ttfamily \footnotesize Please directly complete the \{FUNCTION\_NAME\} function based on the information above. \\
\ttfamily \footnotesize Completed Code: \\
\bottomrule
\end{tabular}
\caption{System prompt for the post-test, where the \texttt{\{PRIOR\_KNOWLEDGE\}} is determined by different levels of simulated students, as described in Table~\ref{tab:low_leve_student},  Table~\ref{tab:mid_leve_student}, and Table~\ref{tab:high_leve_student}.
}
\label{tab:posttest_template}
\end{table}

\paragraph{Knowledge Tracing}
\label{appendix:prompt_KT}

The knowledge tracing in our proposed \model is implemented following the instruction template shown in Table \ref{tab:KT_template}.

\begin{table}[th!]
\renewcommand{\arraystretch}{0.95}
\begin{tabular}{p{0.98\linewidth}}
\toprule
\ttfamily \footnotesize You are a college tutor specializing in Python programming. Your role is to assess a student's understanding of the knowledge required to complete the following task. \\
\ttfamily \footnotesize \\
\ttfamily \footnotesize Task Details: \\
\ttfamily \footnotesize You are guiding a student to complete the  \{FUNCTION\_NAME\}  function from a repository: \\
\ttfamily \footnotesize \verb|```|Python \\
\ttfamily \footnotesize \{TARGET\_CODING\_TASK\} \\
\ttfamily \footnotesize \verb|```| \\
\ttfamily \footnotesize  Reference Knowledge Components (KCs): \\
\ttfamily \footnotesize The dependency paths for the \{FUNCTION\_NAME\} function: \\
\ttfamily \footnotesize \{REFERENCE\_DEPENDENCIES\} \\
\ttfamily \footnotesize - The reference key solution Steps: \\
\ttfamily \footnotesize \{REFERENCE\_STEPS\} \\
\ttfamily \footnotesize \\
\ttfamily \footnotesize Dialogue Context: \\
\ttfamily \footnotesize Below is the ongoing dialogue between you and the student during this tutoring session: \\
\ttfamily \footnotesize \{DIALOGUE\_CONTEXT\} \\
\ttfamily \footnotesize \\
\ttfamily \footnotesize Previous Estimation of Student's Knowledge: \\
\ttfamily \footnotesize \{PREVIOUS\_TURN\_ESTIMATION\} \\
\ttfamily \footnotesize \\
\ttfamily \footnotesize Your Task: \\
\ttfamily \footnotesize Evaluate the student's understanding of the required knowledge components (KCs) based on the dialogue context and previous estimation: \\
\ttfamily \footnotesize - For each KC, determine whether the student has demonstrated understanding or if there is insufficient evidence of understanding. \\
\ttfamily \footnotesize - Mark a KC as ``Known'' if the dialogue provides evidence that the student understands it; mark a KC as ``Unknown'' if there is no evidence in the dialogue that the student understands it. \\
\ttfamily \footnotesize  \\
\ttfamily \footnotesize Output Format: \\
\ttfamily \footnotesize  Provide your evaluation in the following format: \\
\ttfamily \footnotesize - Known knowledge components: [..., ...] \\
\ttfamily \footnotesize - Unknown knowledge components: [..., ...]  \\
\bottomrule
\end{tabular}
\caption{Prompting template for the knowledge tracing in \model.}
\label{tab:KT_template}
\end{table}

\end{document}